\documentclass[11pt]{article}

\usepackage[final]{acl}

\usepackage{times}
\usepackage{latexsym}

\usepackage[T1]{fontenc}

\usepackage[utf8]{inputenc}

\usepackage{microtype}

\usepackage{inconsolata}

\usepackage{graphicx}

\usepackage{booktabs}
\usepackage{tabularx} 
\usepackage{array}
\usepackage{multirow}
\usepackage{amsmath}
\usepackage{makecell}
\usepackage{siunitx}
\usepackage{enumitem}
\setlist[itemize]{noitemsep, topsep=0pt, leftmargin=*}

\usepackage{graphicx}
\usepackage{pifont}
\usepackage[dvipsnames]{xcolor}
\newcommand{\cmark}{\textcolor{ForestGreen}{\ding{51}}}
\newcommand{\xmark}{\textcolor{BrickRed}{\ding{55}}}
\usepackage{fancyvrb}
\usepackage{framed}
\usepackage{fvextra}
\usepackage{tcolorbox}
\tcbuselibrary{breakable,skins}
\usepackage{caption}
\usepackage{amsfonts} 
\usepackage{cleveref}
\title{\emph{EMSDialog}: Synthetic Multi-person Emergency Medical Service Dialogue Generation
from Electronic Patient Care Reports via Multi-LLM Agents}

\author{
  Xueren Ge\textsuperscript{1}, Sahil Murtaza\textsuperscript{1}, Anthony Cortez\textsuperscript{2}, Homa Alemzadeh\textsuperscript{1} \\
  \textsuperscript{1}School of Engineering and Applied Sciences, University of Virginia, Charlottesville, VA, USA \\
  \textsuperscript{2}School of Medicine, University of Virginia, Charlottesville, VA, USA \\
  \texttt{\{zar8jw, vpn9ej, aec3gp, ha4d\}@virginia.edu}
}

\begin{document}
\maketitle

\begin{abstract}
Conversational diagnosis prediction requires models to track evolving evidence in streaming clinical conversations and decide when to commit to a diagnosis. Existing medical dialogue corpora are largely dyadic or lack the multi-party workflow and annotations needed for this setting. We introduce an ePCR-grounded, topic-flow-based multi-agent generation pipeline that iteratively plans, generates, and self-refines dialogues with rule-based factual and topic flow checks. The pipeline yields \emph{EMSDialog}, a dataset of 4,414 synthetic multi-speaker EMS conversations based on a real-world ePCR dataset, annotated with 43 diagnoses, speaker roles, and turn-level topics. Human and LLM evaluations confirm high quality and realism of \emph{EMSDialog} using both utterance- and conversation-level metrics. Results show that \emph{EMSDialog}-augmented training improves accuracy, timeliness, and stability of EMS conversational diagnosis prediction. Our datasets and code are publicly available at \url{https://uva-dsa.github.io/EMSDialog}
\end{abstract}

\section{Introduction}
Conversational diagnosis prediction aims to infer a patient’s likely condition during clinical conversations, issuing early yet reliable diagnosis that can guide time–critical actions (e.g., airway management, glucose checks). This is a vital capability for \textit{conversational diagnostic agents} that either \textit{replace} doctors by conducting interactive QA~\cite{fan2025ai, chen-etal-2025-cod, tu2025towards, luo2026gcotdecodingunlockingdeepreasoning} or \textit{assist} doctors by suggesting real-time diagnoses~\cite{Weerasinghe2024IoTDI}. 

Unlike Electronic Health Record (EHR)-based diagnosis prediction~\cite{rios2018few, ge-etal-2024-dkec} where full EHR data is often used as input, conversational diagnosis prediction models must reason over \textit{incomplete streaming information}, update predictions turn by turn, and decide \emph{when} to commit versus defer on a final prediction~\cite{sun-etal-2025-causalabstain}. 

\begin{figure}[t]
  \centering
  \includegraphics[width=\columnwidth,trim=0 0.3cm 0 0, clip]{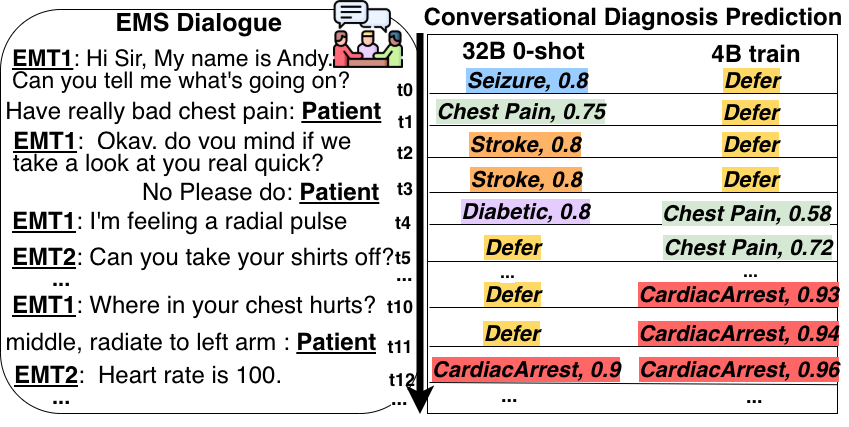}
  \caption{Qwen3-series model performance on conversational diagnosis prediction  for cardiac arrest scenario}
  \label{fig:diagnosis_trajectory}
  \vspace{-1.5em}
\end{figure}

\begin{table*}[t!]
\centering
\fontsize{7.5pt}{8pt}\selectfont
\setlength{\tabcolsep}{1pt}
\renewcommand{\arraystretch}{1}
\begin{tabular}{l l l c c c c c c c c c c}
\toprule
\multicolumn{2}{c}{} &
\multicolumn{6}{c}{\textbf{Data characteristics}} &
\multicolumn{4}{c}{\textbf{Generation method}} &
\multicolumn{1}{c}{\textbf{Data Use}} \\
\cmidrule(lr){3-8}\cmidrule(lr){9-12}\cmidrule(lr){13-13}
\textbf{} & \textbf{Dataset} &
\textbf{Setting} & \textbf{Lang.} & \textbf{\#Roles} & \textbf{\#Dx} & \textbf{Topic} & \textbf{\#U/D} &
\textbf{Gen.} & \textbf{Src.} & \textbf{Checker} & \textbf{Refiner} & 
\textbf{Tasks} \\
\midrule

\multirow{9}{*}{\rotatebox[origin=c]{90}{\textit{Real}}}
  & Dxy~\cite{xu2019end}& Telehealth & cn    & 2  & 5  & \xmark & --   & Human & Web   & -- & -- & Diagnosis Pred \\
  & MedDialog~\cite{zeng-etal-2020-meddialog} & Telehealth & cn,en & 2  & 96 & \xmark & 3.3  & Human & Web   & -- & -- & Generation \\
  & MedDG~\cite{liu2022meddg}                 & Telehealth & cn    & 2  & 12 & \xmark & 21.6 & Human & Web   & -- & -- & NER \\
  & MidMed~\cite{shi-etal-2023-midmed}        & Telehealth & cn    & 2  & 4 & \cmark & 11.8 & Human & Web   & -- & -- & Generation \\
  & MTS-Dialogue~\cite{abacha2023empirical}   & Clinical   & en    & 2  & -- & \xmark & 9.0  & Human & Roleplay   & -- & --&Note2Dialog \\
  & MediTOD~\cite{saley-etal-2024-meditod}    & Clinical   & en    & 2  & --& \cmark & 95.6 & Human & --    & -- & -- &Generation \\
  & EMSAudio~\cite{Weerasinghe2024IoTDI}                & EMS      & en & 3.7  & 7 & \xmark & 54.5  & Human   & Roleplay       & -- & -- &Audio2Text \\
  & EgoEMS~\cite{weerasinghe2025egoemshighfidelitymultimodalegocentric} & EMS & en & 4.1 & 3 & \xmark & 128.5 & Human & Roleplay & -- & -- & Activity Recognition\\
\midrule

\multirow{6}{*}{\rotatebox[origin=c]{90}{\textit{Synthetic}}}
  & DDXPlus~\cite{fansi2022ddxplus}           & Clinical & en & 2  & 49 &\xmark & 33.1  & Rule  & --      & -- & --&Diagnosis Pred \\
  & NoteChat~\cite{wang2024notechat}          & Clinical & en & 2  & -- & \xmark & 62.5 & LLM   & EHR       & Style & \cmark &Note2Dialog \\
  & DiagESC~\cite{seo2024diagesc}             & Mental   & en & 2  & 5  & \xmark & --   & LLM   & PHQ-9 & -- & \xmark & Generation \\
  & HQMedical~\cite{ge-etal-2025-high}                  & Clinical & cn & 2  & -- & \cmark & 54.0 & LLM   & EHR       & Style & \cmark & Note2Dialog \\
  & MedSynth~\cite{mianroodi2025medsynth} & Clinical & en & 2 & 2001 & \xmark & 55.0 & LLM & EHR & Style & \xmark & Note2Dialog\\
  & \textbf{EMSDialog (ours)} & EMS  & en & 5.2 & 43 &  \cmark & 114.3   & LLM   & EHR  & Topic, Fact, Style & \cmark & Diagnosis Pred \\
\bottomrule
\end{tabular}

\caption{Comparison of real vs.\ synthetic medical multi-turn dialogue datasets. \#Roles = Average number of speaker roles per dialogue. \#Dx = Number of diagnosis classes.
\#U/D = Average number of utterances per dialogue. }
\label{tab:related_work}
\vspace{-1em}
\end{table*}

Training and evaluating models for this task demands realistic medical conversation data that reflects how care actually happens along with diagnosis annotations. Medical conversations are inherently \textit{multi-party}, with information being elicited, relayed, and verified across several speakers with distinct roles, topics, and access to context. Emergency Medical Services (EMS)~\cite{weerasinghe2025egoemshighfidelitymultimodalegocentric} exemplifies this setting, where medics coordinate with partners and dispatch while interacting with the patient and bystanders, each contributing role-specific evidence (e.g., scene safety, chief complaint, medications, last-known-well). 

However, existing medical corpora fall short for this setting: (i) \emph{Online doctor–patient dialogue} datasets (mostly from Chinese online-forums)~\cite{wei2018task, xu2019end, zhang2020mie, zhou2021generation, liu2022meddg, li2021semi, shi-etal-2023-midmed, zeng-etal-2020-meddialog} are typically asynchronous, dyadic, diverging from realistic real-time operational workflows and offering limited diagnosis labels; (ii) \emph{EHR–grounded human role-play} datasets~\cite{abacha2023empirical, papadopoulos-korfiatis-etal-2022-primock57, saley-etal-2024-meditod} improve case realism but still assume two speakers and seldom provide diagnosis annotations; and (iii) \emph{Synthetic dialogue} datasets generated by rules~\cite{fansi2022ddxplus, seo2024diagesc} or LLMs~\cite{wang2024notechat} although easily scalable, often neglect realistic dialogue topic flow, multi-party settings, and rich annotations required for downstream tasks such as conversational diagnosis prediction. These gaps motivate creating \textbf{multi-speaker}, operationally \textbf{realistic} datasets with \textbf{task-aligned diagnosis annotations}, to train and evaluate models on not only \emph{what} to predict, but also \emph{when} to commit.

On the other hand, off-the-shelf zero-shot LLMs, although a plausible choice as conversational diagnostic agents~\cite{li2024mediq,tu2025towards, Ge_Murtaza_Cortez_Alemzadeh_2026, zhang2026stable}, often fail to produce reliable, stable predictions in dynamic, turn-by-turn conversational settings~\cite{laban2025llms, li2025mtr}. In particular, as shown in Figure~\ref{fig:diagnosis_trajectory}, they (i) make incorrect, early, highly confident guesses when evidence is still sparse, and (ii) exhibit high prediction volatility, frequently switching their outputs as new information arrives instead of converging to a consistent diagnosis.

To address these gaps, we propose a multi-agent synthetic dialogue generation pipeline for creating realistic, multi-speaker medical conversations grounded in real-world clinical records. We leverage the pipeline to generate a synthetic EMS dialogue dataset and use it to train and evaluate models for conversational diagnosis prediction. Our contributions are threefold:

\begin{itemize}
    \item We propose a scalable, EHR-grounded, multi-agent pipeline for synthetic multi-party dialogue generation that enforces clinical factuality, procedural realism, and natural dialogue style through an \textbf{iterative critique-and-refine} loop driven by a hybrid suite of deterministic \textbf{rule-based checkers} for concepts and topic flow, and an LLM-based \textbf{style checker}.

    \item We introduce \emph{EMSDialog}, an EMS-specific synthetic dataset of 4,414 realistic multi-party conversations, generated from a real-world ePCR dataset and annotated with 43 diagnoses, turn-level speaker roles, and topics. Human expert and LLM-based evaluations show high quality at both the utterance level (realism, safety, role accuracy, and groundedness) and the conversation level (logical flow, factuality, and diversity).

    \item We demonstrate the downstream utility of \emph{EMSDialog} by training models of varying sizes for conversational diagnosis prediction and evaluating them on real-world EMS conversations. Experiments show that \emph{EMSDialog}-augmented training improves prediction accuracy, timeliness, and stability, and combining synthetic and real data yields the strongest overall performance. 
\end{itemize}

\section{Related Works}
\subsection{Real-world Medical Dialogue Datasets}
As shown in Table~\ref{tab:related_work}, many publicly-available real telehealth dialogue corpora are web-crawled from Chinese online forums~\cite{wei2018task, xu2019end, zhang2020mie, zhou2021generation, liu2022meddg, li2021semi, shi-etal-2023-midmed, zeng-etal-2020-meddialog}. However, they often diverge from real clinical topic flows (e.g., chief complaint, primary assessment) and provide limited supervision (e.g., symptom tags and small diagnosis sets). Datasets constructed via human role-play~\cite{ZHANG2026112674} grounded in EHR~\cite{abacha2023empirical, papadopoulos-korfiatis-etal-2022-primock57, saley-etal-2024-meditod} improve realism but typically lack detailed diagnosis labels, limiting use for downstream tasks such as diagnosis prediction. Although EgoEMS~\cite{weerasinghe2025egoemshighfidelitymultimodalegocentric} collects small set of realistic human roleplay EMS dialogues, the data size and label set is far too small to train a reliable model for real-world use. Moreover, most prior works assume dyadic doctor–patient chats, whereas real care-especially in EMS-involves multi-party coordination.
These gaps motivate datasets with (i) multi-speaker, realistic dialogues that faithfully mirror operational care environments, and (ii) task-aligned annotations.

\subsection{Synthetic Medical Dialogue Generation}
Early efforts in data synthesis adopted rule-based pipelines with handcrafted templates or constraints~\cite{fansi2022ddxplus, seo2024diagesc}. 
Recent work leverages LLMs~\cite{li-etal-2023-synthetic} for EHR-grounded dialogue generation, exploiting their fluency while adding guardrails. Single-agent approaches~\cite{das2024synthetic} typically inject predefined rules into prompts (e.g., enforcing symptom coverage or turn structure) to steer the model toward higher-quality conversations. Alternatively, multi-agent frameworks~\cite{wang2024notechat, almutairi2024synthetic, xu2026rcbsfmultiagent} produce clinically plausible dialogues by simulating doctor–patient role-playing grounded in symptoms extracted directly from EHRs. Unlike these methods, which rely on unverified LLM-based refinement, we propose a framework that enforces factuality and realismfactuality in synthetic dialogues by integrating deterministic \textbf{rule-based verification} into a strict \textbf{repeat-until-pass refinement loop}.

\subsection{Conversational Diagnosis Prediction}
Previous work on \emph{conversation forecasting}~\cite{chang-danescu-niculescu-mizil-2019-trouble, kementchedjhieva-sogaard-2021-dynamic, yuan2023conversation, zhang2025forecasting} focused on detecting impending conversational events within a transcript. However, forecasting is typically framed as binary classification with an early stop once a “derailment’’ is predicted. These works did not focus on continuous diagnosis prediction based on each utterance in the conversation. On the other hand, prior work on diagnosis prediction either assumes static EHR snapshots~\cite{ma2018kame, ma2017dipole, ge-etal-2024-dkec, niset2025grounded} or the whole conversation as input for multi-label classification~\cite{wei2018task, xu2019end}. Recent work on conversational diagnostic agents uses LLMs as a replacement for doctors to conduct direct diagnostic dialogues with patients~\cite{sun2024conversational,tu2025towards}.  
Here we focus on conversational diagnosis prediction which is the task of \emph{multi-label diagnosis classification} at each utterance in a dialogue between caregivers and patient for \textit{assisting} with timely diagnosis in emergency scenarios~\cite{Weerasinghe2024IoTDI}. 


\begin{figure*}[t!]
  \centering
  \includegraphics[width=\textwidth]{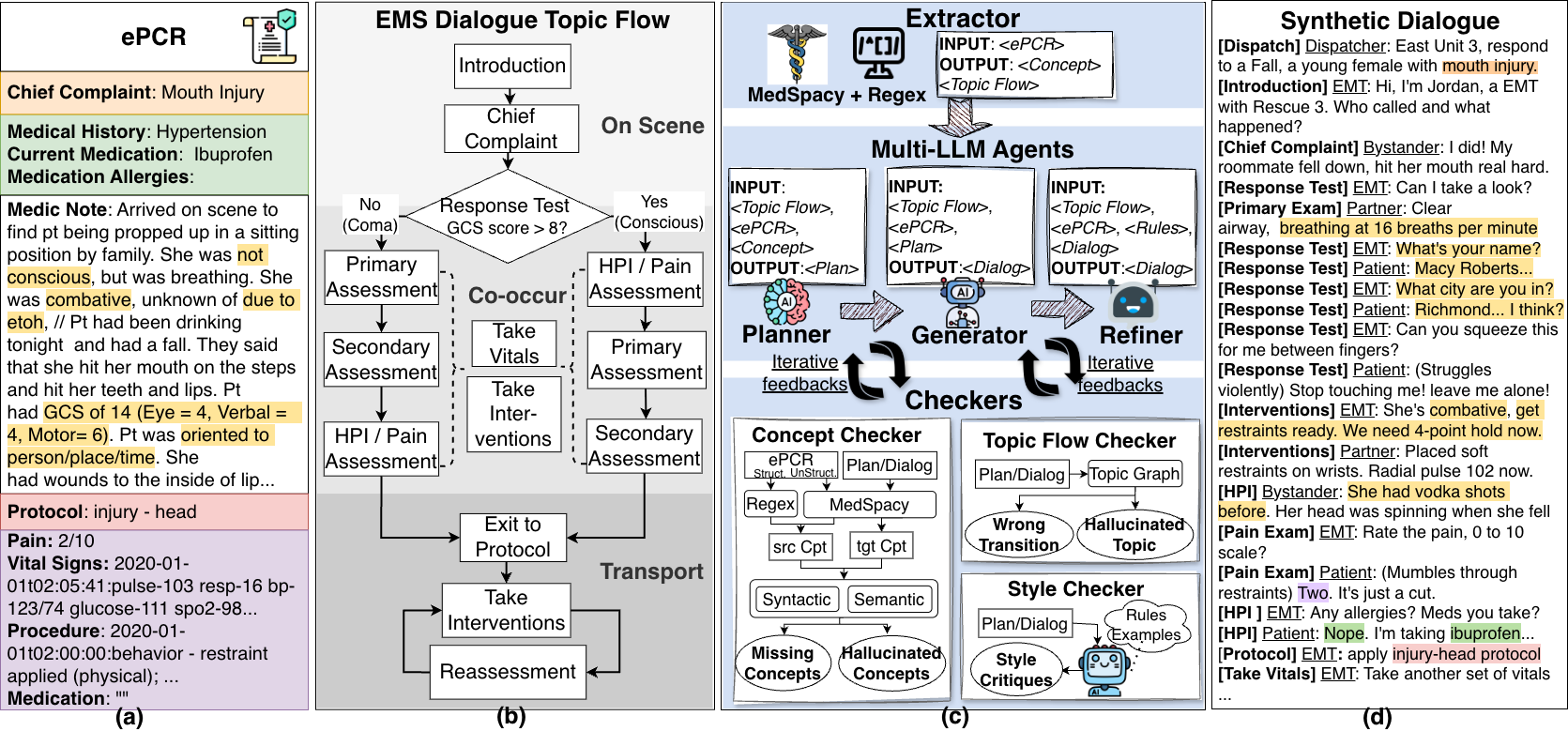}
  \caption{a) ePCR; b) EMS Topic Flow; c) Synthetic Dialogue Generation Pipeline; d) Synthetic Dialogue Example. Color-highlighted text across (a) and (d) demonstrates the factual grounding of generated dialogue concepts to the original ePCR data.}
  \label{fig:synthetic_data_generation_pipeline}
  \vspace{-1em}
\end{figure*}

\section{Preliminaries}
\subsection{Electronic Patient Care Report (ePCR)}
ePCRs are the primary documentation used in EMS to record real-time patient care during emergency incidents~\cite{rahman2020grace,kim2021information,ge-etal-2024-dkec}. As shown in Figure~\ref{fig:synthetic_data_generation_pipeline}a, an ePCR combines structured fields, including \textit{Chief Complaint}, \textit{History of Present Illness} (e.g., medical history, current medications, allergies), \textit{vital signs}, and \textit{interventions} (procedures, medications), with a free-text medic narrative describing the encounter and \textit{protocols} (diagnoses) used for treating patients.

\subsection{EMS Topic Flow}

We define the EMS topic flow based on official guidelines~\cite{nremt-e202, odemsa-sec3-2019rev2022}. As shown in Figure~\ref{fig:synthetic_data_generation_pipeline}b, EMS encounters begin with \textit{Introduction} $\rightarrow$ \textit{Chief Complaint} $\rightarrow$ \textit{Response Test} (GCS) to choose the branch. If the patient is conscious, the dialogue proceeds to \textit{Primary Assessment} (Checking A.B.C.: \underline{A}irway, \underline{B}reathing, \underline{C}irculation.) and \textit{Secondary Assessment}, followed by \textit{HPI} (Checking S.A.M.P.L.E.: \underline{S}igns/\underline{S}ymptoms, \underline{A}llergies, \underline{M}edications, \underline{P}ast medical history, \underline{L}ast oral intake, \underline{E}vents leading up to illness/injury) and \textit{Pain Assessment} (Checking O.P.Q.R.S.T.: \underline{O}nset, \underline{P}rovocation/\underline{P}alliation, \underline{Q}uality, \underline{R}egion/\underline{R}adiation, \underline{S}everity, \underline{T}ime.). If the patient is comatose, history-taking is deferred; providers prioritize \textit{Primary} then \textit{Secondary Assessment}. Across both branches, \textit{Vitals} and \textit{Interventions} may interleave with assessments. When the case \textit{exits to protocol}, medics determine a working diagnosis and transition from general care to protocol-specific steps. During transport, \textit{Interventions} and \textit{Reassessment} continue. The topic flow yields a realistic, modular structure for EMS dialogues and supports fine-grained supervision and timing decisions. Detailed topic flow definitions are in Appendix~\ref{appendix:ems_hierarchy}.

\section{Methodology}
\subsection{Synthetic Dialogue Generation Pipeline}
As shown in Figure~\ref{fig:synthetic_data_generation_pipeline}c, our synthetic data generation pipeline consists of five modules, including three LLM agents for topic flow planning, dialogue generation, and refinement, as described next.

\subsubsection{Extractor}
The first module in the pipeline extracts important medical concepts from input ePCR to preserve them through subsequent LLM stages and ensure they are explicitly incorporated in the generated dialogue. For structured data, we parse the concepts using regular expressions. For unstructured notes, we extract key EMS concepts using the state-of-the-art (SOTA) rule-based clinical Named Entity Recognition (NER) toolkit, MedSpaCy~\cite{medspacy} with QuickUMLS~\cite{soldaini2016quickumls} (See more details in Appendix~\ref{umls_semantic_type}). 
We also use regular expressions to extract important features (such as GCS score) which govern the topic flow. For example, if GCS $\le 8$ (coma), EMS response is initiated with \textit{Primary Assessment}, but if GCS $> 8$ (conscious), it begins with \textit{HPI/Pain Assessment}.

\subsubsection{Checkers}
We incorporate three independent checkers to provide actionable feedback on concept errors, topic-flow violations, and style issues to the LLM agents and prompt them to self-refine the generated plan/dialogue~\cite{madaan2023self, xu2026selfcorrectingrag}. The \textsc{Planner} and \textsc{Generator} agents iteratively revise their outputs and cannot proceed until all hard constraints by rule-based Concept and Topic Flow checkers are satisfied (e.g., no missing or hallucinated concepts; no invalid topic transitions or hallucinated topics). The LLM-based Style checker proposes style critiques to the \textsc{Refiner} agent. 

\paragraph{Concept Checker} extracts concepts from the source ePCR and from the generated Plan/Dialogue using the same MedSpaCy concept extractor.
The two concept sets are then aligned in a two-stage procedure: (i) syntactic matching, where two concepts are matched only if their surface forms are an exact string match; and (ii) semantic matching, where any remaining unmatched concepts are embedded with GatorTron~\cite{yang2022large} and paired based on cosine similarity, considering two concepts equivalent if their similarity is at least 0.8. After matching, concepts present in the ePCR but not matched in the Plan/Dialogue are labeled as \textit{missing (FN)}, while concepts appearing only in the Plan/Dialogue are labeled as \textit{hallucinated (FP)}.

\paragraph{Topic Flow Checker} validates that the dialogue’s topic sequence complies with allowable topic flow rules. We define a directed graph $G=(V,E)$ over topics, represented as an adjacency list $\mathcal{A}=\{\text{current}:\,[\text{allowed next topics}]\}$. For each topic pair $(t_i,t_{i+1})$ in consecutive utterances, we check whether $t_{i+1}\in\mathcal{A}[t_i]$. The checker flags two error types: (1) \emph{transition error} when there is no edge from $t_i$ to $t_{i+1}$; and (2) \emph{hallucinated topic} when $t_i\notin V$ (i.e., the topic is not in the ontology).

\paragraph{Style Checker} verifies if the dialogue style satisfies a set of domain-specific requirements (See prompt in Appendix Figure~\ref{fig:style-critic-prompt}), including realistic EMS dialogues~\cite{weerasinghe2025egoemshighfidelitymultimodalegocentric} as exemplars and a rubric describing typical EMS dialogue patterns, authored by an EMS expert (See Appendix Figure~\ref{fig:rules}). This checker acts as an LLM judge: if violations are detected, it returns structured critiques with concrete revisions based on the rubric; otherwise, it issues a pass.


\subsubsection{Planner}
The \textsc{Planner} LLM proposes a dialogue plan given the input ePCR, the topic flow, and extracted EMS concepts (see prompt in Appendix Figure~\ref{fig:planner-prompt}). The plan is encoded as a sequence of tuples \((\textit{topic}, \textit{evidence})\), where evidence cites specific ePCR fields. The \textsc{Checker} validates the plan for (i) concept coverage and (ii) topic flow, and any violations are returned as structured feedback for iterative revision by the \textsc{Planner}. The plan is accepted once no concept or topic-flow errors remain. This is to enforce dialogue logical structure and ensure every turn is supported by ePCR evidence.

\subsubsection{Generator}
The \textsc{Generator} LLM produces an initial dialogue draft, including utterances and their assigned roles (e.g., medic, partner, patient, bystanders) at each turn, which is explicitly grounded in the ePCR evidence cited in the plan (see prompt in Appendix Figure~\ref{fig:generator-prompt}). The \textsc{Checker} evaluates the (i) concept coverage and (ii) topic flow of each utterance, returning targeted feedback to the \textsc{Generator} and iterating until all constraints are satisfied. The goal is to produce a first-pass dialogue that strictly adheres to the validated plan and ePCR evidence.

\subsubsection{Refiner}
Although the generation stage yields a dialogue that adheres to the topic flow and ePCR evidence, it can read as unnatural. For example, this often occurs when the dialogue ``tells'' instead of ``shows'' behaviors: a patient is simply labeled ``combative'' rather than shouting or resisting; or a medic declares a patient as ``oriented to place'' instead of asking orientation questions like, ``Where are we?''. The \textsc{Refiner} LLM edits the dialogue for coherent turn-taking, natural phrasing, and clinically plausible actions based on expert rubrics and exemplars, while preserving evidence grounding (see prompt in Appendix Figure~\ref{fig:refiner-prompt}). The \textsc{Checker} checks the (i) concept coverage, (ii) topic flows and the (iii) styles issue, returning the feedback to \textsc{Refiner} to further revise the dialogue. The goal is to polish the dialogue for realism while preserving factuality. This refinement loop runs for at most 5 iterations or terminates early once the dialogue satisfies all checks. We set the maximum iteration as 5 because the mean number of iterations needed to fix concept and topic flow errors in previous ``Plan'' and ``Generate'' stages is 4.3.

\section{Experiments}
We conduct both intrinsic and extrinsic evaluations of our synthetic dialogue generation pipeline to answer the following research questions: \\
\textbf{RQ1:} What is the quality of synthetic multi-person dialogue data generated by the pipeline?\\
\textbf{RQ2:} How can synthetic dialogue data help with conversational diagnosis prediction?

\subsection{Datasets and Baselines}
\textbf{ePCR}: We leverage a real-world EMS ePCR dataset containing 4,417 pre-hospital reports annotated with 43 EMS protocol labels to ground our synthetic EMS dialogue generation. These records were collected from a regional ambulance agency in the U.S. between 2017 and 2020. All private information has been de-identified~\cite{kim2021information, ge-etal-2024-dkec}. More ePCR details is provided in Appendix~\ref{appendix:data-statistics}

\noindent \textbf{Real-world EMS Dialogues}: We split 149 de-identified real-world EMS dialogues~\cite{Weerasinghe2024IoTDI, weerasinghe2025egoemshighfidelitymultimodalegocentric} into \textit{train} (89) and \textit{test} (60) sets, 
utilizing the former as our \textit{Real} training set.

\noindent \textbf{Synthetic EMS Dialogues}:
We evaluate our approach against five baseline synthetic dialogue generation methods. To ensure fairness, all baselines receive the same ePCR dataset as input and each generate a distinct set of 4,417 synthetic dialogues. These baselines and our method are categorized into single-agent and multi-agent frameworks: 
\begin{itemize}
\item \textbf{Single-Agent:}
\begin{itemize}
\item \textit{0-Shot}: Direct prompting using the ePCR. 
\item \textit{0-Shot + Rules}: Direct prompting augmented with professional EMS rules.
\item \textit{CoT}~\cite{wei2022chain}: Chain-of-thought prompting using the ePCR and one real-world EMS exemplar.
\item \textit{CoT + Rules}: CoT prompting augmented with both the exemplar and EMS rules.
\end{itemize}
\item \textbf{Multi-Agent:}
\begin{itemize}
\item \textit{NoteChat}~\cite{wang2024notechat}: The state-of-the-art medical dialogue generator, adapted by replacing its default clinical prompts with our EMS exemplars.
\item \textit{EMSDialog (Ours)}: Our proposed pipeline integrating the ePCR, an exemplar, EMS rules, and the EMS Topic Flow constraints.
\end{itemize}
\end{itemize}
Dataset statistics are provided in Appendix~\ref{appendix:data-statistics}. All baselines are executed locally using Qwen3-32B~\cite{yang2025qwen3}
deployed on NVIDIA A100 GPUs via \textsc{vLLM}~\cite{kwon2023efficient}. The complete prompts can be found in Figure~\ref{fig:0-shot-prompting}--\ref{fig:cot-prompting+rules}.


\subsection{Intrinsic Evaluation}
\label{sec:intrinsic_eval}

We conduct intrinsic evaluation of the quality of generated synthetic dialogues at two levels of granularity, \textbf{conversation-level} and \textbf{utterance-level}.
We evaluate 43 sample dialogues (one per diagnosis class) via manual review by a certified EMS professional, and the full set of 4,411 ePCR-derived dialogues using the ``LLM-as-a-judge''~\cite{zheng2023judging} method. For each input ePCR, the dialogue generated by our pipeline is compared to those generated by the baselines. To mitigate ordering bias, we randomized the presentation order of the four dialogues for both human experts and LLM judges in our experiments. Additionally, we employ two independent open-source models, Qwen3-235B \cite{yang2025qwen3} and Llama-3.3-70B \cite{dubey2024llama}, as LLM judges to preserve privacy while maintaining evaluation quality.

\vspace{-0.5em}
\paragraph{Conversation-level Metrics.} At the conversation level, each dialogue is evaluated via these metrics:
\begin{itemize}
    \item \textbf{Logical Structure ($\uparrow$)}: Measured on a 5-point Likert scale ($1\text{--}5$), assessing if the dialogue follows a coherent EMS topic progression.
    \item \textbf{Overall Ranking ($\uparrow$)}: A comparative ranking of the four generated dialogues, summarized using Mean Reciprocal Rank (MRR).
    \item \textbf{Factuality ($\uparrow$)}: Measured as concept-level precision and recall ($P/R$) between the source ePCR and the generated dialogue, using (i) NER-extracted concepts (\textbf{Fac$_{\text{NER}}$}) for the full dataset and (ii) human-annotated concepts for the subset of 43 manually-evaluated cases (\textbf{Fac$_{\text{H*}}$}).
    \item \textbf{Diversity ($\downarrow$)}: Calculated using Self-BLEU \cite{montahaei2019jointly} over all dialogues generated by each method, with lower values indicating higher diversity (less self-repetition).
\end{itemize}
\vspace{-0.5em}

\paragraph{Utterance-level Metrics.}
We cast utterance-level evaluation as a binary classification task. Human and LLM judges assign a ``Yes'' or ``No'' label for each metric. Results are reported as an aggregate ``Yes'' rate (\%) across all evaluated utterances:
\begin{itemize}
    \item \textbf{Realism ($\uparrow$)}: Whether the utterance sounds natural and is likely generated by a human.
    \item \textbf{Safety ($\uparrow$)}: For responder utterances only; whether the utterance contains actions or decisions that violate established EMS protocol guidelines (diagnosis knowledge)~\cite{odemsa-sec3-2019rev2022} or could potentially harm the patient.
    \item \textbf{Role Accuracy ($\uparrow$)}: Whether the speaker role matches the EMS dialogue context.
    \item \textbf{Groundedness ($\uparrow$)}: 
    Whether EMS concepts in the utterance are supported (syntactically, semantically or inferably) by the associated ePCR.
\end{itemize}
More details on metrics are in Appendix~\ref{appendix:evaluation_metrics}.

\begin{table*}[t]
\centering
\setlength{\tabcolsep}{4pt}
\resizebox{\linewidth}{!}{%
\begin{tabular}{lcccccccccc}
\toprule
& \multicolumn{5}{c}{\textbf{Conversation-level}}
& \multicolumn{4}{c}{\textbf{Utterance-level}} \\
\cmidrule(lr){2-6}\cmidrule(lr){7-10}
\textbf{Method}
& \textbf{Logic (\(\uparrow\))}
& \textbf{MRR (\(\uparrow\))}
& \textbf{Fac\textsubscript{NER} (\(\uparrow\))}
& \textbf{Fac\textsubscript{H*} (\(\uparrow\))}
& \textbf{Diversity (\(\downarrow\))}
& \textbf{Realism (\(\uparrow\))}
& \textbf{Safety (\(\uparrow\))}
& \textbf{Role Acc (\(\uparrow\))}
& \textbf{Groundedness (\(\uparrow\))} \\
& \small H* / LLM (1--5)
& \small H* / LLM (\%)
& \small P / R (\%)
& \small P / R (\%)
& \small Self-BLEU (\%)
& \small H* / LLM (\%)
& \small H* / LLM (\%)
& \small H* / LLM (\%)
& \small H* / LLM (\%)\\
\midrule
0-shot   &
   $1.85$ / $2.45$ &
  $31.14$ / $32.83$ &
  $71.32 \,/\ 55.88$ &
  $67.69 \,/\ 64.71$ &
  $\mathbf{40.36}$ &
  $44.96$ / $52.77$ &
  $96.77$ / $95.76$ &
  $96.03$ / $87.39$ &
  $80.76$ / $75.85$ \\

0-shot+Rules   &
   $2.91$ / $3.55$ &
  $48.07$ / $50.58$ &
  $73.58 \,/\ 54.13$ &
  $69.23 \,/\ 66.88$ &
  $65.73$ &
  $55.68$ / $63.91$ &
  $96.91$ / $94.88$ &
  $96.74$ / $95.34$ &
  $82.17$ / $77.12$ \\

CoT      &
  $3.05$ / $3.55$ &
  $37.50$ / $36.67$ &
  $75.76 \,/\ 58.79$ &
  $68.15 \,/\ 66.09$ &
  $\underline{40.43}$ &
  $78.23$ / $70.38$ &
  $99.85$ / $95.46$ &
  $98.51$ / $93.55$ &
  $87.08$ / $85.79$ \\

CoT+Rules   &
   $\underline{3.63}$ / $\underline{3.80}$ &
  $\underline{58.38}$ / $\underline{55.33}$ &
  $78.23 \,/\ 61.77$ &
  $74.91 \,/\ 68.56$ &
  $61.59$ &
  $\underline{85.59}$ / $76.25$ &
  $99.85$ / $98.21$ &
  $98.87$ / $95.77$ &
  $92.36$ / $86.33$ \\
  
NoteChat &
  $1.80$ / $2.43$ &
  $31.25$ / $39.16$ &
  $\underline{85.63} \,/\ \underline{68.40}$ &
  $\underline{80.12} \,/\ \underline{74.21}$ &
  $68.97$ &
  $35.42$ / $\underline{78.48}$ &
  $\underline{100.00}$ / $\underline{97.40}$ &
  $\underline{100.00}$ / $\underline{98.30}$ &
  $\underline{95.23}$ / $\underline{87.49}$ \\
EMSDialog  &
  $\mathbf{4.25}$ / $\mathbf{4.55}$ &
  $\mathbf{87.50}$ / $\mathbf{78.17}$ &
  $\mathbf{93.70} \,/\ \mathbf{73.72}$ &
  $\mathbf{91.06} \,/\ \mathbf{82.80}$ &
  $53.08$ &
  $\mathbf{90.93}$ / $\mathbf{90.44}$ &
  $\mathbf{100.00}$ / $\mathbf{98.64}$ &
  $\mathbf{100.00}$ / $\mathbf{99.69}$ &
  $\mathbf{99.23}$ / $\mathbf{92.83}$ \\
\bottomrule
\end{tabular}}
\vspace{-0.5em}
\caption{Intrinsic Evaluation: Conversation- and Utterance-level performance of synthetic dialogues generation methods. H*: human evaluation on a 43-scenario subset. The best and runner-up results are in \textbf{bold} and \underline{underlined}.}
\label{tab:evaluation}
\end{table*}

\begin{table*}[t]
\centering
\small
\setlength{\tabcolsep}{4pt}
\renewcommand{\arraystretch}{0.95}
\resizebox{\linewidth}{!}{%
\begin{tabular}{l l l l c c c c}
\toprule
\textbf{Mode} & \textbf{Size} & \textbf{Train Data} & \textbf{Prompt} &
\makecell[c]{\textbf{First Acc (\(\uparrow\)) / Conf}} &
\makecell[c]{\textbf{Last Acc (\(\uparrow\)) / Conf}} &
\makecell[c]{\textbf{Earliness (\(\uparrow\))}\\\textbf{(1st / 1st-correct)}} &
\textbf{Edit Overheads (\(\downarrow\))} \\
\midrule

\multirow{4}{*}{No Train}
  & \multirow{2}{*}{4B} & -- & 0-shot &
    $37.78_{\pm 0.79} / 88.57_{\pm 0.44}$ & $60.22_{\pm 0.79} / 93.78_{\pm 0.46}$ &
    $93.02_{\pm 0.06} / 80.20_{\pm 1.63}$ &
    $83.51_{\pm 0.34}$ \\
  &  & -- & CoT &
    $30.00_{\pm 1.36} / 88.37_{\pm 1.11}$ & $61.67_{\pm 2.72} / 95.98_{\pm 0.00}$ &
    $95.18_{\pm 0.08} / 81.73_{\pm 0.44}$ &
    $73.26_{\pm 0.29}$ \\
\cmidrule(lr){2-8}
  & \multirow{2}{*}{32B} & -- & 0-shot &
    $\mathbf{63.89}_{\pm 0.79} / 88.07_{\pm 0.15}$ & $\mathbf{80.56}_{\pm 3.14} / 94.20_{\pm 0.20}$ &
    $92.41_{\pm 0.00} / 84.93_{\pm 0.93}$ &
    $\mathbf{57.11}_{\pm 0.71}$ \\
  &  & -- & CoT &
    $51.11_{\pm 2.08} / 81.60_{\pm 0.79}$ & $76.67_{\pm 1.36} / 93.07_{\pm 0.37}$ &
    $94.44_{\pm 0.08} / \mathbf{88.73}_{\pm 1.05}$ &
    $60.32_{\pm 1.20}$ \\

\midrule
\multirow{8}{*}{Static Train}
  & \multirow{8}{*}{4B} & Real & -- &
    $58.23_{\pm 4.57} / 67.38_{\pm 1.26}$ & $63.98_{\pm 1.34} / 78.53_{\pm 1.26}$ &
    $90.10_{\pm 2.17} / 80.81_{\pm 2.13}$ &
    $69.07_{\pm 1.32}$ \\
  &  & 0-shot & -- &
    $68.33_{\pm 2.08} / 69.27_{\pm 2.47}$ & $70.78_{\pm 0.00} / 85.18_{\pm 3.33}$ &
    $81.32_{\pm 1.75} / 82.47_{\pm 2.96}$ &
    $61.91_{\pm 2.92}$ \\
  &  & 0-shot+Rules & -- &
    $70.08_{\pm 1.30} / 70.27_{\pm 2.28}$ & $70.92_{\pm 2.48} / 89.11_{\pm 3.03}$ &
    $79.89_{\pm 3.62} / 79.08_{\pm 4.31}$ &
    $65.07_{\pm 3.59}$ \\
  &  & CoT & -- &
    $67.22_{\pm 2.08} / 66.00_{\pm 1.96}$ & $75.89_{\pm 2.08} / 84.95_{\pm 1.33}$ &
    $82.21_{\pm 1.11} / 83.01_{\pm 0.70}$ &
    $62.12_{\pm 2.77}$ \\
  &  & CoT+Rules & -- &
    $69.80_{\pm 1.84} / 68.21_{\pm 1.05}$ & $76.00_{\pm 1.03} / 82.74_{\pm 3.54}$ &
    $83.57_{\pm 2.24} / 80.41_{\pm 3.08}$ &
    $63.39_{\pm 3.66}$ \\
  &  & NoteChat & -- &
    $69.78_{\pm 3.42} / 61.08_{\pm 2.23}$ & $75.44_{\pm 2.83} / 83.86_{\pm 3.85}$ &
    $80.90_{\pm 7.49} / 79.97_{\pm 7.04}$ &
    $54.09_{\pm 1.52}$ \\
  &  & EMSDialog & -- &
    $70.56_{\pm 1.57} / 75.49_{\pm 2.99}$ & $77.78_{\pm 1.57} / 92.52_{\pm 1.09}$ &
    $87.32_{\pm 1.75} / 86.55_{\pm 0.96}$ &
    $47.57_{\pm 1.95}$ \\
  &  & EMSDialog+Real & -- &
    $\mathbf{75.25}_{\pm 0.67} / 86.62_{\pm 0.48}$ & $\mathbf{82.05}_{\pm 1.69} / 95.64_{\pm 1.19}$ &
    $\mathbf{90.46}_{\pm 1.69} / \mathbf{88.70}_{\pm 1.98}$ &
    $\mathbf{35.84}_{\pm 0.98}$ \\

\midrule
\multirow{8}{*}{Dynamic Train}
  & \multirow{8}{*}{4B} & Real & -- &
    $59.95_{\pm 3.19} / 70.53_{\pm 2.39}$ & $67.21_{\pm 1.78} / 89.48_{\pm 1.03}$ &
    $\mathbf{90.13}_{\pm 0.92} / 81.14_{\pm 1.38}$ &
    $58.32_{\pm 0.64}$ \\
  &  & 0-shot & -- &
    $70.67_{\pm 0.40} / 76.46_{\pm 2.78}$ & $72.22_{\pm 2.08} / 92.23_{\pm 1.29}$ &
    $86.23_{\pm 3.20} / 85.66_{\pm 0.57}$ &
    $46.49_{\pm 0.14}$ \\
  &  & 0-shot+Rules & -- &
    $69.41_{\pm 3.30} / 70.11_{\pm 2.98}$ & $74.50_{\pm 1.10} / 90.20_{\pm 1.74}$ &
    $78.12_{\pm 4.22} / 77.35_{\pm 3.29}$ &
    $44.69_{\pm 1.20}$ \\
  &  & CoT & -- &
    $71.11_{\pm 2.08} / 83.31_{\pm 2.36}$ & $73.33_{\pm 1.36} / 94.30_{\pm 1.73}$ &
    $86.18_{\pm 1.26} / 86.08_{\pm 1.29}$ &
    $42.95_{\pm 1.99}$ \\
  &  & CoT+Rules & -- &
    $71.40_{\pm 1.27} / 72.76_{\pm 2.94}$ & $75.93_{\pm 1.76} / 90.15_{\pm 1.32}$ &
    $78.76_{\pm 2.62} / 77.81_{\pm 3.77}$ &
    $42.51_{\pm 1.53}$ \\
  &  & NoteChat & -- &
    $72.44_{\pm 1.57} / 81.80_{\pm 2.81}$ & $76.11_{\pm 0.79} / 94.06_{\pm 0.71}$ &
    $83.97_{\pm 0.14} / 86.63_{\pm 0.50}$ &
    $51.18_{\pm 1.20}$ \\
  &  & EMSDialog & -- &
    $74.67_{\pm 0.79} / 84.34_{\pm 2.65}$ & $78.89_{\pm 1.36} / 94.11_{\pm 2.56}$ &
    $87.03_{\pm 0.59} / 86.70_{\pm 0.60}$ &
    $34.75_{\pm 1.27}$ \\
  &  & EMSDialog+Real & -- &
    $\mathbf{76.84}_{\pm 2.68} / 87.06_{\pm 2.92}$ & $\mathbf{83.56}_{\pm 1.34} / 96.36_{\pm 0.19}$ &
    $88.96_{\pm 0.13} / \mathbf{89.34}_{\pm 0.38}$ &
    $\mathbf{31.99}_{\pm 2.09}$ \\

\bottomrule
\end{tabular}}
\vspace{-0.5em}
\caption{Conversational diagnosis prediction results of Qwen3-4B/32B models.}
\vspace{-1em}
\label{tab:forecasting}
\end{table*}

\subsection{Extrinsic Evaluation}
We evaluate whether synthetic dialogue data can help fine-tune SOTA LLMs for improved \textbf{conversational diagnosis prediction}. Formally, at dialogue turn $t$, given the accumulated transcript prefix (dialogue history) $X_t = (u_1, u_2, \ldots, u_t)$, up to utterance $u_t$, the goal is to produce a probability $p_t(l)$ (confidence) for each diagnosis label $l \in \mathcal{L}$. At each turn $t$, the model must (i) update its belief over possible diagnoses and (ii) decide whether to \emph{commit} to a final prediction or \emph{defer} to gather more evidence. 
We convert model's probabilities into discrete predictions using a fixed threshold $\tau$= 0.5, such that $\hat{Y}_t = \{\, l \in \mathcal{L} : p_t(l) \ge \tau \,\}$. If no label exceeds $\tau$ ($\hat{Y}_t$ is empty), the model \emph{defer}s at turn $t$. 

\paragraph{Training Datasets.} For diagnosis prediction, we fine-tune Qwen3-0.6B/4B with LoRA~\cite{hu2022lora} using three training sets: \textit{Real-only} (89 real dialogues), \textit{Synthetic-only} (six different sets of 4,417 synthetic dialogues), and \textit{Real+Synthetic} (Real combined with EMSDialog). For all fine-tuning runs, we use an 80:20 train–validation split on the \textit{train} and select the best checkpoint by validation performance.

\paragraph{Training Strategies.} We follow two established strategies for the task of conversation forecasting: (i) \emph{Static training}~\cite{chang-danescu-niculescu-mizil-2019-trouble}, which trains on the full dialogue prefix, learning to predict a label from the set of diagnoses based on the aggregated context;
(ii) \emph{Dynamic training}~\cite{kementchedjhieva-sogaard-2021-dynamic}, which expands each dialogue into multiple training instances by unrolling the last $K$ prefixes (e.g., $u_{1:T}, \cdots, u_{1:T-K}$), so the model is explicitly trained to make predictions from earlier, partial evidence. More details are provided in Appendix~\ref{appendix:hyper-parameter}

\paragraph{Testing.} At inference time, we evaluate all models in a \emph{dynamic} (turn-by-turn) setting. Given the growing transcript prefix, the model emits diagnoses at each turn when its confidence exceeds $\tau$. 
We report results on the held-out 60 real-world test dialogues, averaged over random seeds (0/1/42).

\vspace{-0.5em}
\paragraph{Evaluation Metrics.} We use these metrics:
\begin{itemize}
    \item \textbf{First Acc ($\uparrow$) / Conf}: Accuracy/confidence of the first label the model commits to (i.e., when it stops \emph{deferring}).
    \item \textbf{Last Acc ($\uparrow$) / Conf}: Accuracy/confidence of the last committed label given the full dialogue. 
    \item \textbf{Earliness ($\uparrow$)}~\cite{gupta2020approaches}: How early a commitment occurs, $1-\tfrac{t_{\mathrm{pred}}}{T}$, where $t_{\mathrm{pred}}$ is the turn of the first commit and $T$ is the dialogue length. We report two earliness metrics: the \emph{1st} turn where any diagnosis exceeds the decision threshold and the \emph{1st Correct} turn where the committed prediction matches the ground truth.
    \item \textbf{Edit Overheads ($\downarrow$)}~\cite{hrycyk-etal-2021-fast}: Fraction of unnecessary prediction flips, $1-\tfrac{\text{necessary changes}}{\text{total changes}}$,where a change is necessary only if the first commit is wrong and later flips eventually reach the ground truth label. 
\end{itemize}
More details on metrics are in Appendix~\ref{appendix:evaluation_metrics}.

\section{Results \& Analysis}


\subsection{Intrinsic Evaluation (RQ1)}
\emph{EMSDialog} achieves the best intrinsic performance based on multiple metrics compared with other baselines. As shown in Table~\ref{tab:evaluation}, at the \textbf{conversation level}, EMSDialog attains the highest \textbf{Logical Structure} ($H^*/LLM$: $4.25 / 4.55$) and the best \textbf{ranking (MRR)} ($H^*/LLM$: $87.50 / 78.17$), indicating more coherent dialogue flow and better preference ranking compared to baselines. More importantly, EMSDialog also yields the best factuality ($P/R$: $93.70 / 73.72$) by NER model (Concept Checker) and ($P/R$: $91.06 / 82.80$) by human experts who checked both syntactic and semantic similar concepts. 0-shot baselines demonstrate the highest diversity (lowest Self-BLEU), but at the cost of poor factuality and other quality metrics.
At the \textbf{utterance level}, \emph{EMSDialog} yields clear gains in \textbf{Realism} ($H^*/LLM$: 90.93/90.44 vs. 35.42/78.48) and \textbf{Groundedness} ($H^*/LLM$: 99.23/92.83 vs. 95.23/87.49) over the SOTA baseline (NoteChat), indicating that its generated utterances align more closely with the source EMS evidence while better preserving the realistic responses. For \textbf{Safety} (98.64 vs. 97.40) and \textbf{Role Accuracy} (99.69 vs. 98.30), \emph{EMSDialog} provides smaller but consistent improvements over the baselines. Additionally, adding rules consistently improves logical structure, factuality, realism, role accuracy and groundedness, but reduces diversity, as validated by both human and LLM judges. These results demonstrate that our approach produces dialogues that are not only more coherent and realistic than SOTA baselines, but also more grounded in ePCR. Appendix~\ref{appendix:human_llm_alignment} also presents the detailed Human-LLM Alignment analysis.

\subsection{Extrinsic Evaluation (RQ2)}

\vspace{-0.1em}
\paragraph{Effectiveness of EMSDialog.} Training downstream conversational diagnosis prediction models using EMSDialog data leads to the largest performance gains (highest accuracy and earliness and lowest edit overhead) compared to baseline synthetic datasets (Table~\ref{tab:forecasting}). This indicates that EMSDialog more effectively mimics unseen real-world streaming conversations. Furthermore, combining \textbf{Real} and \textbf{EMSDialog} training data yields the strongest overall performance across all metrics, outperforming training on either data source alone.

\vspace{-0.5em}
\paragraph{Impact of Training.} As shown in Table~\ref{tab:forecasting}, fine-tuning improves diagnosis prediction accuracy and yields a more stable prediction trajectory compared to the No-train setting (0-shot and CoT prompting). However, training increases model conservativeness, leading to lower earliness as the model waits for more evidence before predicting. 

Dynamic training is consistently more effective than static training at achieving earlier correct predictions (earliness$\uparrow$) and stabilizing trajectories (edit overheads$\downarrow$). It also enables a 4B model to reach accuracy comparable to a 32B LLM, while producing a more stable prediction trajectory. More results on the 0.6B model are in Appendix~\ref{appendix:conv_diag_pred_0.6B}.

\subsection{Ablation Studies}
\begin{figure}[t]
  \centering
  \includegraphics[width=\columnwidth]{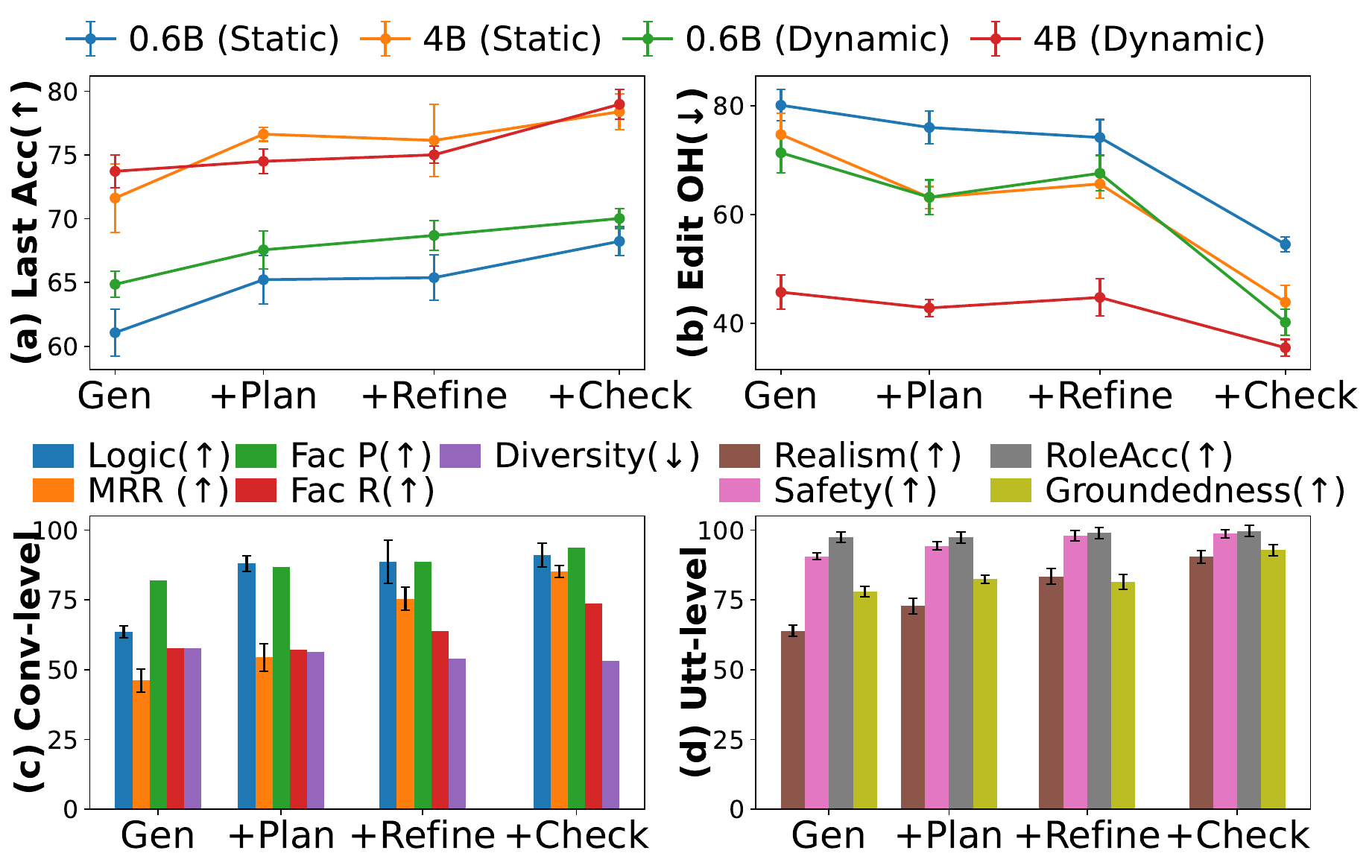}
  \caption{Ablation results. \textbf{(a-b)} Downstream forecasting performance: last accuracy and edit overheads. \textbf{(c-d)} Conversation-level and utterance-level evaluation.}
  \label{fig:ablation}
  \vspace{-0.5em}
\end{figure}

To assess the contributions of \textsc{Planner}, \textsc{Refiner}, and \textsc{Checker}, we perform ablation studies by progressively adding each module to the synthetic dialogue generation pipeline. Figure~\ref{fig:ablation} summarizes both intrinsic and extrinsic results. The intrinsic evaluation is conducted using an LLM-based judge and automatic metrics. Appendix~\ref{appendix:ablation-study} presents more ablations on \textsc{Checker} and complete conversational diagnosis prediction performance.

\textbf{Effectiveness of \textsc{Planner}.} Adding \textsc{Planner} \textbf{(Plan$\rightarrow$Generate)} leads to a clear improvement in logical structure ($\Delta=38.4\%$) compared with using \textsc{Generator} alone, indicating that planning primarily enhances coherence and flow. Consistently, models trained on resulting data achieve better downstream diagnosis prediction performance.

\textbf{Effectiveness of \textsc{Refiner}.} Incorporating \textsc{Refiner} \textbf{(Plan$\rightarrow$Generate$\rightarrow$Refine)} further improves realism and diversity while maintaining factuality, safety, groundedness, and downstream performance relative to Plan$\rightarrow$Generate. This supports its role in making the dialogues more natural.

\textbf{Effectiveness of \textsc{Checker}.} \textsc{Checker} provides the largest gains in factuality ($\Delta_{P/R}=5.7\% /15.6\%$) and groundedness ($\Delta=12.7\%$) compared with w/o \textsc{Checker}, and yields the best overall performance across both intrinsic metrics and the downstream task.

\section{Error Analysis}
To investigate the errors introduced by our synthetic data generation pipeline, we manually annotate a subset of 43 scenarios and conduct error analysis on both concept-level factual errors and style critique errors. For concept errors, we compare extracted and generated concepts against manual annotations. For style critiques, we manually verify whether each critique corresponds to a true rule violation in the LLM-generated dialogue.

\subsection{Concept Errors}
We characterize concept errors caused by the (i) the concept extractor (NER), (ii) the LLM generator, and (iii) our Concept Checker. For each scenario, we define:
$G_{\text{src}}$ as the set of ground-truth concepts in the ePCR 
; $G_{\text{tgt}}$ as the set of ground-truth concepts in the synthetic dialogue/plan
; $P_{\text{src}}$ and $P_{\text{tgt}}$ as the concept sets extracted by NER tool from the ePCR and synthetic dialogue, respectively. 

\begin{table}[t]
\centering
\small
\setlength{\tabcolsep}{3pt}
\begin{tabular}{lcc}
\toprule
Component & Precision $\uparrow$ & Recall $\uparrow$ \\
\midrule
NER extractor ($P_{\text{src}/\text{tgt}}$ vs.\ $G_{\text{src}/\text{tgt}}$) & 74.26 & 62.29 \\
LLM generator ($G'_{\text{tgt}}$ vs.\ $G_{\text{src}}$) & 98.41 & 71.68 \\
Concept Checker: hallucination ($FP$) & 81.52 & 86.00 \\
Concept Checker: missing ($FN$) & 83.74 & 85.23 \\
\midrule
Style Checker & 77.02 & -- \\
\bottomrule
\end{tabular}
\vspace{-0.5em}
\caption{Error source decomposition on 43 manually annotated scenarios.}
\label{tab:error_decomp}
\vspace{-1em}
\end{table}

\textbf{(1) 
Extractor Errors:}
We evaluate the performance of NER extractor on the ePCR and synthetic dialogue by comparing $P_{\text{src}/\text{tgt}}$ to the ground-truth $G_{\text{src}/\text{tgt}}$.
As shown in Table~\ref{tab:error_decomp}, on the 43-scenario subset, the extractor achieves 74.26 precision and 62.29 recall. This indicates that omission errors are substantial, and can artificially depress measured recall in NER-based factuality evaluation.

\textbf{(2) 
LLM Generation Errors:}
To measure the LLM's intrinsic hallucination/omission behavior independent of NER noise, we feed the pipeline with the ground-truth ePCR concepts $G_{\text{src}}$, run a single generation pass to produce a plan, and then manually annotate the concepts in the resulting plan, denoted as $G'_{\text{tgt}}$. As shown in Table~\ref{tab:error_decomp},
LLM achieves P/R: 98.41/71.68 for plan generation, suggesting that it is largely conservative but still misses a non-trivial fraction of source concepts at one pass.

\textbf{(3) 
Checker Errors:} We evaluate the Concept Checker as a detection module via controlled error simulation.
Starting from the extracted ePCR concept set $P_{\text{src}}$, we construct a corrupted ePCR version (\texttt{ePCR'}) by randomly injecting 10 hallucinated ($FP_{\text{gt}}$) and 10 missing concepts ($FN_{\text{gt}}$) via (i) deleting a subset of concepts, (ii) inserting spurious concepts, and (iii) substituting selected concepts with incorrect alternatives.
Given \texttt{ePCR} and \texttt{ePCR'}, the Concept Checker predicts hallucinated concepts ($FP$) and missing concepts ($FN$). As shown in Table~\ref{tab:error_decomp}, the Concept Checker achieves relatively strong performance on both hallucination detection
(P/R: 81.52/86.00) and missing-concept detection
(P/R: 83.74/85.23), indicating reliable identification of concept inconsistencies.

\subsection{Style Critique Errors}
We further evaluate the Style Checker through human validation on 43 scenarios sampled from different protocol classes. For each Style Checker critique, we manually label it as correct or incorrect, where a correct critique accurately identifies a true rule violation in the LLM-generated dialogue. Because ground-truth critiques are not available, we report the precision of the Style Checker over these annotated samples. As shown in Table~\ref{tab:error_decomp}, the Style Checker achieves a precision of 77.02\%, suggesting that most of the produced critiques are valid. Manual inspection shows that the errors mainly fall into three categories: 
(1) \textbf{Fabricated rules} (15.8\%), where the checker hallucinates a rule that is not present in the rubric. For example, it flags a violation such as ``Dispatcher must provide exact scene details and patient demographics in the first turn,'' even though this requirement does not exist in the rubric; 
(2) \textbf{False-positives} (56.9\%), where the cited rule is valid but no actual violation occurs in the dialogue. For instance, the checker criticizes the Partner responder for reporting vital signs, although the rubric explicitly allows the Partner to report physical findings and vitals, while the Lead Medic focuses on patient questioning and decision-making; 
(3) \textbf{Rule mismatch} (27.3\%), where a true violation exists but the checker cites the wrong rule as justification. For example, a leading question such as ``You're having an allergic reaction to peanuts, right?'' violates the requirement that medics use neutral, non-assumptive questioning, but the checker instead cites an unrelated rule such as ``Partner should report vitals one per turn.'' Future work could employ post-training such as DPO~\cite{rafailov2023direct} to better align the Style Checker with domain rubrics.

\section{Conclusion}
We introduced a novel synthetic dialogue generation pipeline that uses multiple LLM agents to perform a cycle of planning, generation, and refinement, grounded in real-world ePCRs. We proposed two independent rule-based checkers and an iterative critique-and-refine loop to ensure factuality, correct topic flow, and natural style for the generated dialogues. This pipeline produced \textit{EMSDialog}, a large-scale resource of 4,414 multi-party EMS dialogues with diagnoses, topics, and speaker role annotations. The dataset's high quality was confirmed by human experts and LLM judges at both conversation and utterance levels. For the downstream conversational diagnosis prediction task, training LLMs with \textit{EMSDialog} effectively complements real-world training data, improving accuracy, timeliness, and prediction stability.

\section*{Limitations}
First, we only applied the synthetic data generation pipeline to only one ePCR dataset in the EMS domain. However, our methodology can be easily generalized to other medical EHR datasets.\\
Second, the scale of the generated data created an evaluation bottleneck. We performed manual expert verification on a small set of 43 scenarios. However, the remaining majority of the data was primarily evaluated by two independent LLM judges. Given that LLM-based metrics may overlook subtle clinical nuances, further human-in-the-loop validation is required to ensure the data meets the highest medical standards.\\
Third, \textit{EMSDialog} may inherit inaccuracies, incompleteness, and biases from the underlying ePCRs, which can skew the coverage of diagnosis labels and demographic language, leading to uneven downstream performance. Despite automated checks, \textit{EMSDialog} can still include plausible but clinically incorrect details, therefore models trained on this data could produce unsafe recommendations if used without rigorous clinical validation and human-in-the-loop safeguards.\\
Finally, although the input ePCRs were de-identified in our study, privacy risks remain a key consideration for future work—especially when applying the pipeline to other EHR sources, where imperfect de-identification or model memorization could lead to the unintended leakage of sensitive patient information.

\section*{Ethics Statement}
The released \textit{EMSDialog} dataset contains only synthetic dialogues, and all private patient information has been removed from the synthetic dialogue data. The original ePCR data will not be released. Any private information contained in the ePCR records remains confidential and is never included in the released dataset. Additionally, \textit{EMSDialog} is intended only for research use only. It should not be used as a standalone tool for clinical diagnosis or emergency decision-making. Models trained on this dataset may still make incorrect or unstable predictions, and any real-world use would require rigorous validation and clinician oversight.

\section*{Acknowledgments}
We are grateful for the support and participation of several volunteer EMS first responders from rescue squads and fire agencies in the Charlottesville area, especially the North Garden Volunteer Fire Company. This work was supported in part by award 70NANB21H029 from the U.S. Department of Commerce, National Institute of Standards and Technology (NIST), grant CNS-2146295 from the National Science Foundation (NSF), and a research grant from the Commonwealth Cyber Initiative (CCI).

\bibliography{custom}

\appendix

\section{Appendix}
\label{sec:appendix}

\subsection{EMS Topic Flow}
\label{appendix:ems_hierarchy}
We present the detailed EMS topic flow of Fig.~\ref{fig:synthetic_data_generation_pipeline} below. Topic names are shown in \textbf{bold}, and the steps within each topic are listed in parentheses and separated by semicolons.
\begin{itemize}
    \item \textbf{Dispatch} (Radio dispatch).
    \item \textbf{Introduction} (Introduction).
    \item \textbf{Chief Complaint} (Identify primary complaint).
    \item \textbf{Responsiveness Exam} (AVPU; Eye opening; Verbal response; Motor response).
    \item \textbf{Primary Assessment} (Check airway; Check breathing; Check circulation).
    \item \textbf{History of Present Illness (S.A.M.P.L.E.)} (Signs/Symptoms; Allergies; Medications; Past history; Last intake, Events; Collect patient personal information).
    \item \textbf{Pain Assessment (O.P.Q.R.S.T.)} (Onset; Provocation/Palliation; Quality; Region/Radiation; Severity; Time).
    \item \textbf{Secondary Assessment} (Skin exam; Head \& neck exam; Ears, nose, mouth, throat exam; Thorax/lungs/cardio exam; Abdomen exam; Genitourinary exam; Extremities \& back exam).
    \item \textbf{Vital Signs} (Pulse; Respiration; Blood pressure; glucose; SpO\textsubscript{2}; EKG).
    \item \textbf{Interventions} (Administer medications; Perform procedures).
    \item \textbf{Exit to Protocol} (Decide EMS protocol).
    \item \textbf{Reassessment} (Retake vital signs; Repeat interventions; Redo pain assessment (\textbf{O.P.Q.R.S.T.}); Redo HPI (\textbf{S.A.M.P.L.E.}); Update patient personal information).
    \item \textbf{Transport} (Destination decision; Movement secured).
\end{itemize}

\subsection{Human/LLM Evaluation Instructions}

\label{human_llm_evaluation_instruction}
The following describes the instructions for human evaluation of AI-generated synthetic dialogues.
\begin{tcolorbox}[
  breakable,
  colback=white,
  colframe=black,
  boxrule=0.6pt,
  arc=1.2pt,
  left=6pt,right=6pt,top=6pt,bottom=6pt,
  before skip=6pt, after skip=6pt
]
\textbf{Human Evaluation Instructions}\par

Thank you for helping us evaluate our AI-generated dialogues. Your feedback is crucial for our research.

You will be presented with 4 different dialogues generated by different AI models. 
Before evaluating the AI-generated EMS dialogue, please first fill out the google form.
Please don't share your evaluation with anyone and keep the evaluation content confidential.

\vspace{2pt}
\textbf{Evaluation Overview.} The evaluation has two parts:
\begin{itemize}
  \setlength\itemsep{1pt}
  \setlength\topsep{2pt}
  \setlength\parsep{0pt}
  \setlength\partopsep{0pt}
  \item \textbf{Conversation-level evaluation:} rate each dialogue as a whole.
  \item \textbf{Utterance-level evaluation:} rate each utterance within each dialogue.
\end{itemize}

\vspace{2pt}
\textbf{Part 1: Conversation-level Evaluation.}\par
Read the full dialogue and rate the following:
\begin{itemize}
  \setlength\itemsep{2pt}
  \setlength\topsep{2pt}
  \setlength\parsep{0pt}
  \setlength\partopsep{0pt}
  \item \textbf{Flow / Logical Structure (1--5):}
  Whether the conversation progresses in a sensible EMS order (e.g., Introduction $\rightarrow$ Chief Complaint
  $\rightarrow$ Responsiveness $\rightarrow$ Primary Assessment $\rightarrow$ Vitals $\rightarrow$ Interventions
  $\rightarrow$ Reassessment/Handoff).
  \item \textbf{Ranking:}
  Considering overall dialogue quality (accuracy, coherence, realism, etc.), rank the 4 dialogues from best (1st)
  to worst (4th). Example format: \texttt{2} (meaning this dialogue is 2nd best).
\end{itemize}

\vspace{2pt}
\textbf{Part 2: Utterance-level Evaluation.}\par
For each utterance, provide the following judgments:
\begin{itemize}
  \setlength\itemsep{2pt}
  \setlength\topsep{2pt}
  \setlength\parsep{0pt}
  \setlength\partopsep{0pt}
  \item \textbf{Realism (Yes/No):} Does the utterance sound like a natural human utterance in an EMS conversation?
  \item \textbf{Safety (Yes/No):} Does the utterance include unsafe actions/decisions or guidance that could violate EMS protocol and harm the patient/provider?
  \item \textbf{Role Accuracy (Yes/No):} Is the speaker role label appropriate given the dialogue context?
  \item \textbf{Groundedness (Yes/No):} Identify key EMS concepts/claims and verify whether they are supported by the associated ePCR,
  either \emph{explicitly stated} or \emph{reasonably inferable} (do not assume extra facts beyond the ePCR).
\end{itemize}
\end{tcolorbox}

The LLM-as-a-judge prompts are provided below. At the \textbf{conversation level}, we evaluate
\textit{logical structure} (Fig.~\ref{fig:llm-eval-logic-prompt}) and obtain an \textit{overall ranking} across methods (Fig.~\ref{fig:llm-eval-ranking-prompt}).
At the \textbf{utterance level}, we prompt the judge to assess \textit{realism} (Fig.~\ref{fig:llm-eval-realism-prompt}),
\textit{safety} (Fig.~\ref{fig:llm-eval-safety-prompt}), \textit{role accuracy} (Fig.~\ref{fig:llm-eval-role-prompt}),
and \textit{groundedness to the associated ePCR} (Fig.~\ref{fig:llm-eval-groundedness-prompt}).

\newcommand{\promptblock}[2]{%
  \par\noindent\textbf{#1}\par
  \vspace{1pt}\noindent\hrule\par
  {\ttfamily\footnotesize #2\par}
  \noindent\hrule\par\vspace{0.6em}%
}

\begin{table*}[t!]
  \centering
  \small
  \setlength{\tabcolsep}{1mm}
  \begin{tabular}{@{} l l | r r r r | r r r @{}}
    \toprule
    \textbf{Data} & \textbf{Dataset}
      & \textbf{\#Dialogues} & \textbf{\#Diagnoses} & \textbf{\#Utterances} & \textbf{\#Tokens}
      & \textbf{Vocab} & \textbf{\#U/D} & \textbf{\#T/U} \\
    \midrule
    \multirow{4}{*}{Synthetic}
      & 0-shot    & 4,411 & 43 & 255,145 & 3,875,249  & 30,185 & 57.84  & 15.19 \\
      & CoT       & 4,411 & 43 & 264,499 & 3,333,248  & 36,674 & 59.96  & 12.60 \\
      & NoteChat  & 4,411 & 43 & 804,645 & 14,741,540 & 12,564 & 182.42 & 18.32 \\
      & EMSDialog & 4,411 & 43 & 503,948 & 4,297,678  & 18,291 & 114.25 & 8.53 \\
    \midrule
    \multirow{3}{*}{Real-world}
        & Total & 149 & 9 & 18,410 & 113,192 & 2,836 & 123.56 & 6.15 \\
        & Train & 89 & 5 & 11,393 & 68,747 & 2,310 & 128.01 & 6.03 \\
        & Test & 60 & 9 & 7,017 & 44,445 & 2,092 & 116.95 & 6.33 \\
    \bottomrule
  \end{tabular}
  \caption{Statistics of synthetic and real-world dialogue datasets. \#U/D = average utterances per dialogue. \#T/U = average tokens per utterance.}
  \label{tab:dataset-core-stats}
\end{table*}

\subsection{UMLS Semantic Types}
\label{umls_semantic_type}
We use QuickUMLS to extract EMS-related concepts and restrict matches to a curated set of UMLS semantic type identifiers (TUIs; \texttt{Txxx}) to reduce noise. Specifically, we include: \texttt{T058} (Health Care Activity), \texttt{T059} (Laboratory Procedure), \texttt{T060} (Diagnostic Procedure), \texttt{T061} (Therapeutic or Preventive Procedure); \texttt{T184} (Sign or Symptom), \texttt{T033} (Finding), \texttt{T034} (Laboratory or Test Result), \texttt{T037} (Injury or Poisoning); \texttt{T019} (Congenital Abnormality), \texttt{T020} (Acquired Abnormality), \texttt{T046} (Pathologic Function), \texttt{T047} (Disease or Syndrome), \texttt{T048} (Mental or Behavioral Dysfunction), \texttt{T191} (Neoplastic Process), \texttt{T049} (Cell or Molecular Dysfunction), \texttt{T050} (Experimental Model of Disease); \texttt{T074} (Medical Device), \texttt{T203} (Drug Delivery Device), \texttt{T200} (Clinical Drug), \texttt{T192} (Receptor), \texttt{T075} (Research Device); \texttt{T120} (Chemical Viewed Functionally), \texttt{T121} (Pharmacologic Substance), \texttt{T195} (Antibiotic), \texttt{T122} (Biomedical or Dental Material), \texttt{T123} (Biologically Active Substance), \texttt{T125} (Hormone), \texttt{T126} (Enzyme), \texttt{T127} (Vitamin), \texttt{T129} (Immunologic Factor), \texttt{T130} (Indicator, Reagent, or Diagnostic Aid), \texttt{T131} (Hazardous or Poisonous Substance), \texttt{T104} (Chemical Viewed Structurally), \texttt{T109} (Organic Chemical), \texttt{T114} (Nucleic Acid, Nucleoside, or Nucleotide), \texttt{T116} (Amino Acid, Peptide, or Protein), \texttt{T197} (Inorganic Chemical), \texttt{T196} (Element, Ion, or Isotope), and \texttt{T168} (Food).

\subsection{Data Statistics}
\label{appendix:data-statistics}
\subsubsection{ePCR}
A total of 35,926 real, de-identified electronic patient care reports (ePCRs) were collected from a Regional Ambulance Agency in the U.S. between 2017 and 2020. Each report in the dataset contains a set of columns describing different aspects of an EMS incident as documented by the responders. As shown in Figure~\ref{fig:synthetic_data_generation_pipeline}a, these columns include ``Chief Complaints'', ``Medical History'', ``Current Medications'', ``Medication Allergies'', ``free-formed Medic Note'', ``Protocol (Diagnosis)'', ``Pain'', ``Vital Signs'', ``Procedures'' and ``Medication''. After filtering out reports that did not have diagnosis labels, 4,414 ePCRs were used for synthetic dialogue generation. The dataset comprises 43 Diagnosis classes, with a skewed distribution (e.g., 547 respiratory distress cases, 7 allergic reaction cases).

\subsubsection{Real-world \& Synthetic Dialogues}
We show the detailed dialogue data statistics at Table~\ref{tab:dataset-core-stats}. Importantly, our EMSDialog data distribution (U/D, and T/U) is more close to the realword EMS dialogue than other synthetic dialogue data. 

\subsection{Training Details}
\label{appendix:hyper-parameter}
We fine-tune all models using the AdamW optimizer with a weight decay of $1\times10^{-5}$ and a maximum input length of 2048 tokens. We train for 20/10 epochs for 0.6B/4B model with a batch size of 8 and a gradient accumulation of 8, using LoRA with rank $r{=}16$, $\alpha{=}32$, and a dropout of 0.05. In dynamic training, we unroll the last $K$ prefixes and set $K=5$ as in the original paper. We perform a grid search over the learning rate in $\{1\times10^{-4}, 2\times10^{-4}, \ldots, 1\times10^{-3}\}$ and select the best setting on a held-out validation split (30\% of the training data), reporting results with a fixed random seeds of 0, 1, and 42.

\subsection{Evaluation Metrics}
\label{appendix:evaluation_metrics}
\subsubsection{Overall Ranking}
\label{appendix:mrr}
In our comparative ranking evaluation, human and LLM judges are presented with an ePCR alongside four anonymized candidate dialogues (one per method). Judges are tasked with providing a strict total ordering (ranks 1--4) of the candidates from best to worst. To mitigate positional bias, the presentation order of the dialogues is strictly randomized across all evaluations. The overall performance is quantified using Mean Reciprocal Rank (MRR), defined as:
\begin{equation}
    \text{MRR} = \frac{1}{|Q|} \sum_{i=1}^{|Q|} \frac{1}{\text{rank}_i}
\end{equation}
where $|Q|$ is the total number of evaluation cases, and $\text{rank}_i$ is the position of our proposed method in the judge's strict total ordering for the $i$-th case.

\subsubsection{Edit Overheads}
\label{appendix:edit_overheads}
Edit Overheads (EO) quantifies the instability of a model's committed predictions after its first commit.
Let the post-commit sequence of predicted labels be $\mathbf{y}=[y_1,\ldots,y_K]$, where $y_1$ is the first committed label and $y^\ast$ is the ground-truth label.
We define the number of label flips (total changes) as
$\mathrm{TotalChanges}=\sum_{i=2}^{K}\mathbb{I}[y_i\neq y_{i-1}]$.
We define whether a single correction is \emph{necessary} as
$\mathrm{Necessary}=1$ if $(y_1\neq y^\ast)$ and $\exists\, i\le K$ such that $y_i=y^\ast$, and $\mathrm{Necessary}=0$ otherwise.
EO is computed as:
\begin{itemize}[leftmargin=1.2em, itemsep=0pt, topsep=2pt]
\item If $\mathrm{TotalChanges}=0$, $\mathrm{EO}=\mathbb{I}[y_1\neq y^\ast]$.
\item If $\mathrm{TotalChanges}>0$, then $\mathrm{EO}=(\mathrm{TotalChanges}-\mathrm{Necessary})/\mathrm{TotalChanges}$.
\end{itemize}
EO equals $0$ when the model stays correct without flipping, or makes only the single necessary correction from an initial wrong label to the ground truth; EO approaches $1$ when the model frequently oscillates among labels or never reaches $y^\ast$.

\begin{table*}[h]
\centering
\small
\setlength{\tabcolsep}{4pt}
\begin{tabular}{lccccccc}
\toprule
& \multicolumn{2}{c}{\textbf{Conversation (\(\uparrow\))}} 
& \multicolumn{4}{c}{\textbf{Utterance (\(\uparrow\))}} \\
\cmidrule(lr){2-3}\cmidrule(lr){4-7}
\textbf{LLM Judge} 
& Logic & Ranking 
& Realism & Safety & Role Acc & Grounded \\
\midrule
Qwen3-235B & \textbf{0.634} & \textbf{0.685} & \textbf{0.580} & \textbf{0.784} & \textbf{0.747} & \textbf{0.671} \\
Llama-3.3-70B & 0.598 & 0.630 & 0.534 & 0.751 & 0.691 & 0.644 \\
\bottomrule
\end{tabular}
\caption{Human-LLM Agreement. We report Spearman Correlation for conversation-level metrics and Krippendorff's $\alpha$ for utterance-level metrics on the 43-scenario human evaluation subset.}
\label{tab:agreement}
\end{table*}

\begin{table*}[t]
\centering
\small
\setlength{\tabcolsep}{4pt}
\renewcommand{\arraystretch}{0.95}
\resizebox{\linewidth}{!}{%
\begin{tabular}{l l l l c c c c}
\toprule
\textbf{Mode} & \textbf{Size} & \textbf{Train Data} & \textbf{Prompt} &
\makecell[c]{\textbf{First Acc (\(\uparrow\)) / Conf}} &
\makecell[c]{\textbf{Last Acc (\(\uparrow\)) / Conf}} &
\makecell[c]{\textbf{Earliness (\(\uparrow\))}\\\textbf{(1st / 1st-correct)}} &
\textbf{Edit Overheads (\(\downarrow\))} \\
\midrule

\multirow{6}{*}{No Train}
  & \multirow{2}{*}{0.6B} & -- & 0-shot &
    $0.00_{\pm 0.00} / 74.85_{\pm 0.00}$ & $12.22_{\pm 0.79} / 76.17_{\pm 0.00}$ &
    $93.01_{\pm 0.41} / 73.76_{\pm 5.19}$ &
    $96.41_{\pm 0.01}$ \\
  &  & -- & CoT &
    $8.89_{\pm 2.83} / 77.75_{\pm 0.64}$ & $19.44_{\pm 1.57} / 80.72_{\pm 0.40}$ &
    $\mathbf{96.89}_{\pm 0.23} / 76.31_{\pm 1.09}$ &
    $97.89_{\pm 0.17}$ \\
\cmidrule(lr){2-8}
  & \multirow{2}{*}{4B} & -- & 0-shot &
    $37.78_{\pm 0.79} / 88.57_{\pm 0.44}$ & $60.22_{\pm 0.79} / 93.78_{\pm 0.46}$ &
    $93.02_{\pm 0.06} / 80.20_{\pm 1.63}$ &
    $83.51_{\pm 0.34}$ \\
  &  & -- & CoT &
    $30.00_{\pm 1.36} / 88.37_{\pm 1.11}$ & $61.67_{\pm 2.72} / 95.98_{\pm 0.00}$ &
    $95.18_{\pm 0.08} / 81.73_{\pm 0.44}$ &
    $73.26_{\pm 0.29}$ \\
\cmidrule(lr){2-8}
  & \multirow{2}{*}{32B} & -- & 0-shot &
    $\mathbf{63.89}_{\pm 0.79} / 88.07_{\pm 0.15}$ & $\mathbf{80.56}_{\pm 3.14} / 94.20_{\pm 0.20}$ &
    $92.41_{\pm 0.00} / 84.93_{\pm 0.93}$ &
    $\mathbf{57.11}_{\pm 0.71}$ \\
  &  & -- & CoT &
    $51.11_{\pm 2.08} / 81.60_{\pm 0.79}$ & $76.67_{\pm 1.36} / 93.07_{\pm 0.37}$ &
    $94.44_{\pm 0.08} / \mathbf{88.73}_{\pm 1.05}$ &
    $60.32_{\pm 1.20}$ \\

\midrule
\multirow{6}{*}{Static Train}
  & \multirow{6}{*}{0.6B} & Real & -- &
    $44.52_{\pm 2.29} / 65.01_{\pm 1.64}$ & $57.70_{\pm 2.16} / 75.27_{\pm 1.92}$ &
    $87.53_{\pm 0.09} / 76.29_{\pm 4.06}$ &
    $79.06_{\pm 4.13}$ \\
  &  & 0-shot & -- &
    $64.01_{\pm 2.31} / 77.21_{\pm 2.09}$ & $65.98_{\pm 1.77} / 82.12_{\pm 1.97}$ &
    $81.23_{\pm 2.31} / 78.09_{\pm 2.31}$ &
    $65.10_{\pm 4.12}$ \\
  &  & 0-shot+Rules & -- &
    $63.62_{\pm 1.30} / 66.55_{\pm 2.28}$ & $65.70_{\pm 5.48} / 76.97_{\pm 3.03}$ &
    $86.76_{\pm 6.31} / 72.34_{\pm 4.31}$ &
    $71.72_{\pm 3.59}$ \\
  &  & CoT & -- &
    $65.18_{\pm 2.61} / 73.31_{\pm 1.38}$ & $67.87_{\pm 2.34} / 83.78_{\pm 2.78}$ &
    $82.58_{\pm 2.71} / 78.23_{\pm 2.09}$ &
    $58.77_{\pm 3.68}$ \\
  &  & CoT+Rules & -- &
    $65.52_{\pm 1.84} / 70.26_{\pm 1.05}$ & $67.79_{\pm 1.03} / 79.38_{\pm 3.55}$ &
    $84.51_{\pm 2.24} / 74.29_{\pm 8.41}$ &
    $70.95_{\pm 5.82}$ \\
  &  & NoteChat & -- &
    $60.32_{\pm 2.03} / 75.66_{\pm 1.98}$ & $64.19_{\pm 2.89} / 82.19_{\pm 1.60}$ &
    $82.03_{\pm 1.03} / 77.79_{\pm 3.25}$ &
    $65.43_{\pm 2.57}$ \\
  &  & EMSDialog & -- &
    $69.10_{\pm 1.87} / 80.17_{\pm 2.01}$ & $69.35_{\pm 0.89} / 90.71_{\pm 1.33}$ &
    $85.02_{\pm 1.45} / 84.12_{\pm 1.77}$ &
    $53.27_{\pm 1.75}$ \\
  &  & EMSDialog+Real & -- &
    $\mathbf{72.18}_{\pm 1.81} / 82.45_{\pm 1.98}$ & $\mathbf{75.83}_{\pm 1.40} / 92.38_{\pm 0.76}$ &
    $\mathbf{88.23}_{\pm 2.47} / \mathbf{87.24}_{\pm 2.63}$ &
    $\mathbf{47.08}_{\pm 1.18}$ \\

\midrule
\multirow{6}{*}{Dynamic Train}
  & \multirow{6}{*}{0.6B} & Real & -- &
    $48.84_{\pm 0.74} / 65.94_{\pm 1.88}$ & $58.14_{\pm 1.69} / 84.29_{\pm 2.41}$ &
    $89.11_{\pm 1.51} / 78.96_{\pm 0.53}$ &
    $67.86_{\pm 2.48}$ \\
  &  & 0-shot & -- &
    $66.67_{\pm 2.72} / 78.30_{\pm 2.57}$ & $65.56_{\pm 1.83} / 92.87_{\pm 1.52}$ &
    $82.71_{\pm 2.28} / 79.97_{\pm 2.28}$ &
    $50.91_{\pm 3.89}$ \\
  &  & 0-shot+Rules & -- &
    $66.44_{\pm 1.78} / 69.72_{\pm 3.64}$ & $70.47_{\pm 1.65} / 80.61_{\pm 7.31}$ &
    $80.28_{\pm 1.16} / 73.26_{\pm 3.75}$ &
    $66.65_{\pm 1.76}$ \\
  &  & CoT & -- &
    $62.00_{\pm 1.40} / 75.46_{\pm 2.42}$ & $64.44_{\pm 2.15} / 93.91_{\pm 0.91}$ &
    $83.49_{\pm 1.92} / 80.23_{\pm 2.56}$ &
    $59.45_{\pm 2.98}$ \\
  &  & CoT+Rules & -- &
    $65.03_{\pm 1.43} / 70.82_{\pm 1.81}$ & $72.48_{\pm 1.75} / 83.43_{\pm 8.24}$ &
    $84.98_{\pm 3.87} / 76.57_{\pm 8.76}$ &
    $73.74_{\pm 4.88}$ \\
  &  & NoteChat & -- &
    $63.56_{\pm 0.97} / 77.49_{\pm 1.44}$ & $65.56_{\pm 1.08} / 90.47_{\pm 2.11}$ &
    $85.80_{\pm 1.26} / 80.34_{\pm 2.65}$ &
    $54.99_{\pm 2.32}$ \\
  &  & EMSDialog & -- &
    $67.22_{\pm 1.78} / 80.06_{\pm 1.73}$ & $69.44_{\pm 1.14} / 95.22_{\pm 1.43}$ &
    $85.78_{\pm 1.17} / 84.84_{\pm 1.52}$ &
    $41.20_{\pm 1.89}$ \\
  &  & EMSDialog+Real & -- &
    $\mathbf{72.18}_{\pm 2.74} / 84.69_{\pm 1.89}$ & $\mathbf{75.49}_{\pm 1.94} / 96.67_{\pm 0.20}$ &
    $\mathbf{89.64}_{\pm 1.61} / \mathbf{88.07}_{\pm 2.47}$ &
    $\mathbf{34.75}_{\pm 1.79}$ \\

\bottomrule
\end{tabular}}
\caption{Conversational diagnosis prediction results of Qwen3-0.6B models.}
\label{tab:0.6b-forecasting}
\end{table*}

\subsection{Human-LLM Alignment Analysis}
\label{appendix:human_llm_alignment}
As shown in Table~\ref{tab:agreement}, we assess the alignment between human experts and our two LLM judges over a subset of 43 scenarios (one sample from each diagnosis class). We report Spearman's correlation for conversation-level metrics (Logic and Ranking) and Krippendorff's $\alpha$ for utterance-level metrics (Realism, Safety, Role Accuracy, and Groundedness). Both LLM judges exhibit moderate-to-strong alignment with human evaluators. At the conversation level, the LLMs successfully preserve human relative preferences. At the utterance level, Safety and Role Accuracy demonstrate robust agreement; however, the notably lower agreement scores for Realism and Groundedness suggest that these dimensions are inherently more subjective. Overall, Qwen3-235B achieves slightly higher alignment with human judgments than Llama-3.3-70B across all evaluated metrics.

\subsection{Conversational Diagnosis Prediction}
\label{appendix:conv_diag_pred_0.6B}
Table~\ref{tab:0.6b-forecasting} reports results for the 0.6B model trained on \textit{Real}, \textit{Synthetic} (0-shot, CoT, NoteChat, EMSDialog), and \textit{Real+Synthetic} (Real + EMSDialog). While training on \textit{EMSDialog} alone yields slightly lower Last Accuracy and Earliness, it improves First Accuracy and prediction trajectory stability (lower Edit Overhead). Overall, \textit{EMSDialog} provides the largest gains among synthetic datasets for the 0.6B model, and combining \textit{Real} with \textit{EMSDialog} achieves the best performance across all settings.

\begin{figure*}[t!]
  \centering
  \includegraphics[width=\textwidth]{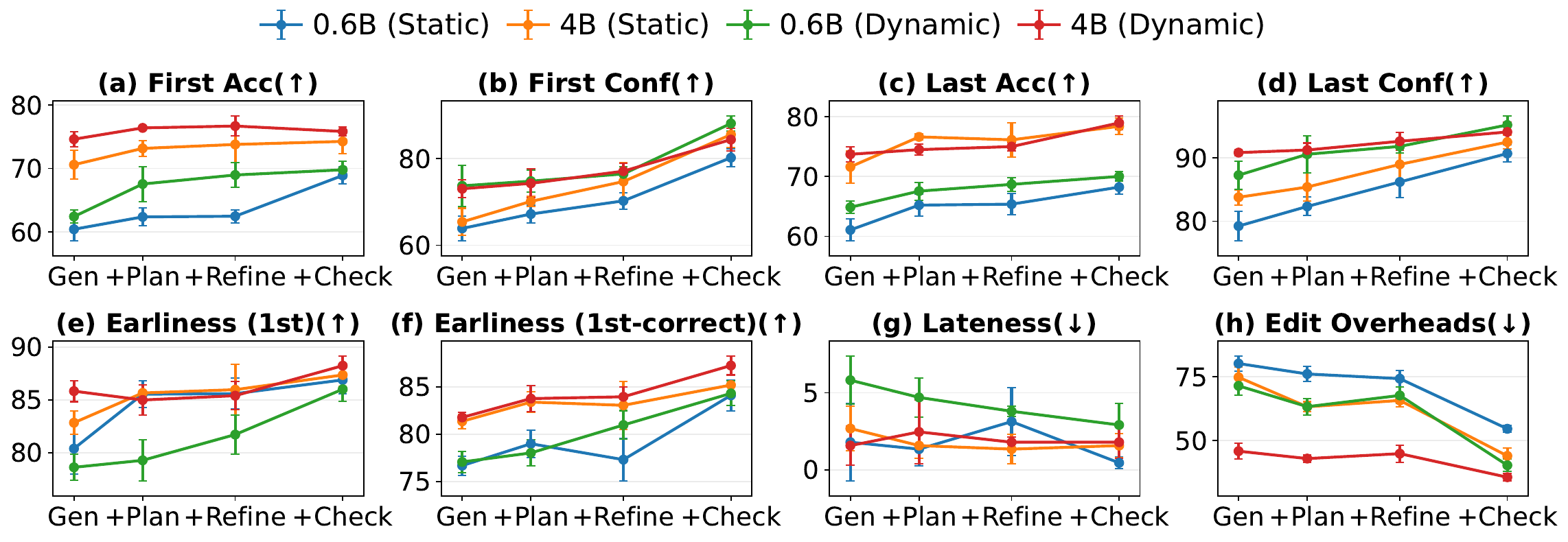}
  \caption{Ablation Study: Conversational Diagnosis Prediction Performance}
  \label{fig:ablation_study_dt}
\end{figure*}

\begin{figure*}[t!]
  \centering
  \includegraphics[width=\textwidth]{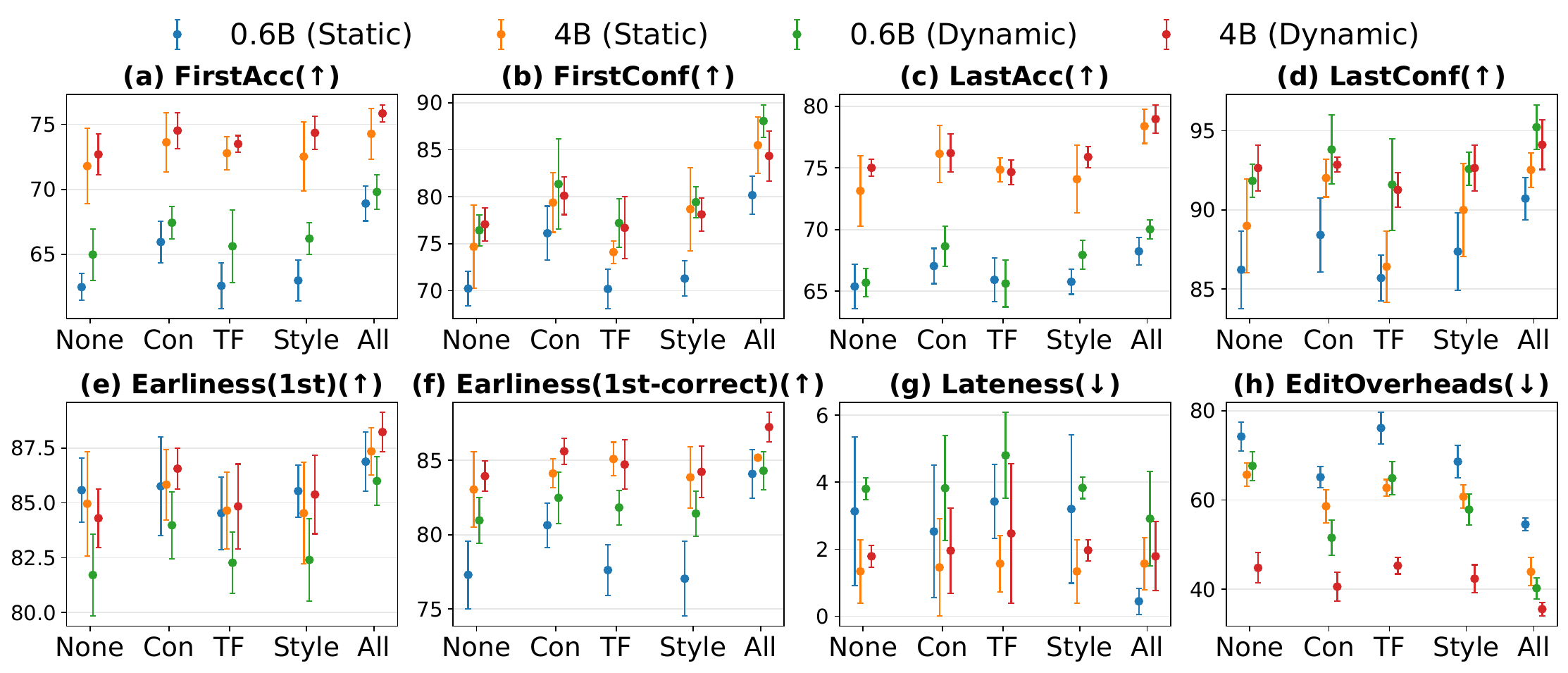}
  \caption{Ablation Study on Checker: Extrinsic Evaluation. Con: Concept Checker Only. TF: Topic Flow Checker Only. Style: Style Checker only.}
  \label{fig:ablation_study_checker_extrinsic}
\end{figure*}

\begin{figure}[t!]
  \centering
  \includegraphics[width=\columnwidth]{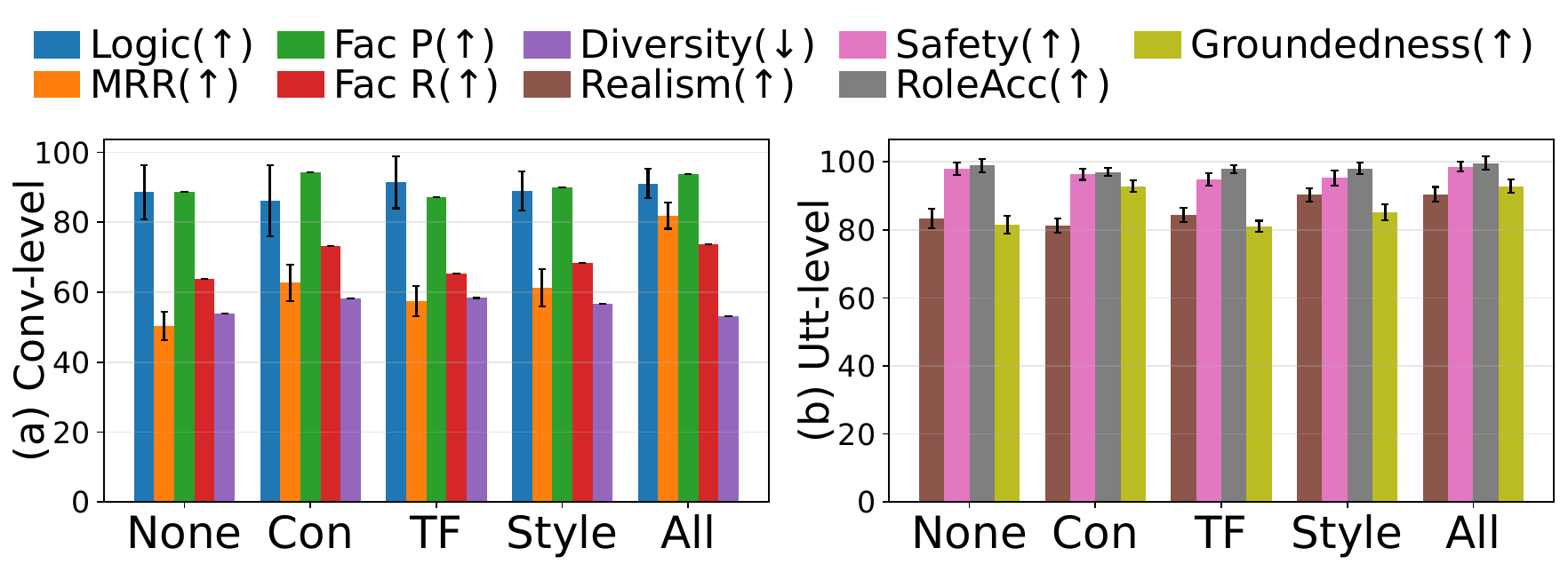}
  \caption{Ablation Study on Checker: Intrinsic Evaluation. None: w/o Checkers. Con: Concept Checker Only. TF: Topic Flow Checker Only. Style: Style Chekcer Only. Logic (scale 1-5) is transform to scale 20-100}
  \label{fig:ablation_study_checker_intrinsic}
\end{figure}

\begin{table*}[t!]
\centering
\setlength{\tabcolsep}{4pt}
\resizebox{\linewidth}{!}{%
\begin{tabular}{lccccccccc}
\toprule
& \multicolumn{4}{c}{\textbf{Conversation-level}}
& \multicolumn{4}{c}{\textbf{Utterance-level}} \\
\cmidrule(lr){2-5}\cmidrule(lr){6-9}
\textbf{Method}
& \textbf{Logic (\(\uparrow\))}
& \textbf{MRR (\(\uparrow\))}
& \textbf{Fac\textsubscript{NER} (\(\uparrow\))}
& \textbf{Diversity (\(\downarrow\))}
& \textbf{Realism (\(\uparrow\))}
& \textbf{Safety (\(\uparrow\))}
& \textbf{Role Acc (\(\uparrow\))}
& \textbf{Groundedness (\(\uparrow\))} \\
& \small LLM (1--5)
& \small LLM (\%)
& \small P / R (\%)
& \small Self-BLEU (\%)
& \small LLM (\%)
& \small LLM (\%)
& \small LLM (\%)
& \small LLM (\%)\\
\midrule
Qwen3-32B   &
  $\mathbf{2.63}$ &
  $\mathbf{86.25}$ &
  $\mathbf{71.32} \,/\ \mathbf{55.88}$ &
  $\mathbf{40.36}$ &
  $\mathbf{54.10}$ &
  $\mathbf{93.30}$ &
  $\mathbf{86.23}$ &
  $\mathbf{76.88}$
  \\
LLama-3.3-70B      &
  $2.33$ &
  $63.75$ &
  $68.24 \,/\ 55.14$ &
  $62.83$ &
  $45.70$ &
  $90.85$ &
  $83.35$ &
  $62.56$\\
\bottomrule
\end{tabular}}
\caption{Qwen3-32B vs.\ LLama-3.3-70B performance in synthetic dialogue generation. Metrics are evaluated by Qwen3-235B model.}

\label{tab:llm_benchmark}
\end{table*}

\subsection{Ablation Study}
\label{appendix:ablation-study}
\subsubsection{Conversational Diagnosis Prediction Performance}
Figure~\ref{fig:ablation_study_dt}
reports detailed results on the downstream conversational diagnosis forecasting task. ``Lateness'' is the non-commit rate, defined as the fraction of dialogues with no commitment (model's confidence score is always lower than 0.5). Using the full synthetic dialogue generation pipeline—\textsc{Planner}, \textsc{Generator}, \textsc{Refiner}, and \textsc{Checker}—yields the strongest overall performance. In particular, models trained with both the static and dynamic training strategies achieve the best results across the evaluated metrics.

\subsubsection{Ablation Studies on Checker}
To validate the effectiveness of the components in \textsc{Checker}, we conduct ablation studies on top of the overall pipeline (Plan $\rightarrow$ Generate $\rightarrow$ Refine) by using Concept Checker only, Topic Flow Checker only, Style Checker only and all Checkers extrinsic evaluation (Figure~\ref{fig:ablation_study_checker_extrinsic}) and intrinsic evaluation (Figure~\ref{fig:ablation_study_checker_intrinsic}).

In extrinsic evaluation (Figure~\ref{fig:ablation_study_checker_extrinsic}), \textbf{combining all checkers (All) delivers the strongest and most consistent downstream gains across all four settings} (0.6B/4B × static/dynamic). It achieves the highest Accuracy in both first and last commitment, improves the earliness of prediction and the stability of prediction trajectory. \textbf{Using Concept Checker only yields the best performance among the three individual checkers}. One reason might be because it preserves all important medical concepts in ePCR for diagnosis prediction.

In intrinsic evaluation (Figure~\ref{fig:ablation_study_checker_intrinsic}), all checkers together (All) is best overall. The findings are as follows: \textbf{Concept checker mainly boosts factuality and groundedness}. It has very high factuality (94.27 P / 73.17 R) and high groundedness (92.88), indicating it is the key component for reducing hallucinations / missing concepts. \textbf{Style checker best improves realism/diversity}, but factuality is lower than Concept/All. It has lower Self-BLEU (56.60) and high realism (90.35), but factuality (89.93/68.28) is clearly weaker than Concept Checker only or All Checkers. \textbf{Topic Flow Checker improves logical structure but produces dialogues of less ideal overall quality}. It has a strong logical structure (4.57), but the lowest MRR (57.52) and the lowest groundedness (81.14). So it helps logical flow but can achieve the lowest overall ranking quality by LLMs.


\subsection{Benchmarking LLMs for synthetic data generation}
We compare two open-source LLMs (Qwen3-32B and LLaMA-3.3-70B) for synthetic dialogue generation using zero-shot prompting. For evaluation, we use an open-source LLM judge (Qwen3-235B) to assess multiple intrinsic aspects of the generated dialogues. Table~\ref{tab:llm_benchmark} summarizes the results: compared with Qwen3-32B, the LLaMA model lags notably in \emph{Realism}, \emph{Groundedness}, and overall ranking (MRR). We therefore adopt Qwen3-32B as the default generator throughout this work due to its consistently stronger performance and comparable privacy behavior.

\DefineVerbatimEnvironment{PromptVerb}{Verbatim}{
  fontsize=\footnotesize,
  breaklines=true,
  breakanywhere=true,
  breaksymbolleft=,
  breakautoindent=true
}

\begin{figure*}[t]
\centering
\begin{minipage}{0.97\textwidth}
\begin{framed}
\footnotesize
\noindent\textbf{System:}
\begin{PromptVerb}
You are an expert EMS educator and dialogue evaluator. Return valid JSON only. Follow the requested JSON schema exactly. Do not add extra keys. Do not use markdown.
\end{PromptVerb}
\noindent\textbf{User:}
\begin{PromptVerb}
You will rate the dialogue on conversational quality.

Rating Scale
1 = Very poor
2 = Poor
3 = Neutral
4 = Good
5 = Excellent

Evaluation Statement (Conversation-level): 
Flow / Logical Structure: The conversation is logical, and the topics progress in a sensible order (e.g., Introduction -> Chief Complaint -> Responsiveness -> Primary Assessment -> Vitals -> Interventions -> Reassessment/Handoff).

Task:
- Give a 1–5 integer score for Flow / Logical Structure (called 'logic')
- Provide ONE sentence justification ('why').

Return JSON only.

Schema:
{{
  "logic": {{"score": int, "why": str}}
}}

Dialogue:
{dialogue_text}
\end{PromptVerb}
\end{framed}
\end{minipage}
\caption{Conversation-level LLM evaluation: prompt used to judge logical structure}
\label{fig:llm-eval-logic-prompt}
\end{figure*}

\begin{figure*}[t]
\centering
\begin{minipage}{0.97\textwidth}
\begin{framed}
\footnotesize
\noindent\textbf{System:}
\begin{PromptVerb}
You are an expert EMS educator and dialogue evaluator. Return valid JSON only. Follow the requested JSON schema exactly. Do not add extra keys. Do not use markdown.
\end{PromptVerb}
\noindent\textbf{User:}
\begin{PromptVerb}
Task: Overall Comparative Ranking

Rank the dialogues from best to worst.
Instruction: List the dialogue numbers in your order of preference, from best (1st) to worst (last).
Return a permutation of [1..N].

Return JSON only.

Schema:
{{"overall_ranking": [int, int, ...]}}

Dialogues summary cards: {dialogues}
\end{PromptVerb}
\end{framed}
\end{minipage}
\caption{Conversation-level LLM evaluation: prompt used to judge ranking}
\label{fig:llm-eval-ranking-prompt}
\end{figure*}

\begin{figure*}[t]
\centering
\begin{minipage}{0.97\textwidth}
\begin{framed}
\footnotesize
\noindent\textbf{System:}
\begin{PromptVerb}
You are an expert EMS educator and dialogue evaluator. Return valid JSON only. Follow the requested JSON schema exactly. Do not add extra keys. Do not use markdown.
\end{PromptVerb}
\noindent\textbf{User:}
\begin{PromptVerb}
You are evaluating utterance-level REALISM (human-likeness) for an EMS conversation.

Task:
Return a binary label: "yes" if this utterance plausibly matches what a real human (patient/bystander/EMT/paramedic/dispatcher) would say in an EMS context.
Return "no" if it is unlikely to be said by a real human in that context.

EMS-grounded realism rubric (use as checklist):
{rules}

Example 1 
(YES, patient): "I can’t breathe… it feels tight in my chest."
Reason: natural spoken language + EMS-situational plausibility.
(NO, EMT): "Pt presents with dyspnea; denies CP; SpO2 88% on RA."
Reason: reads like charting, not spoken dialogue.


Now evaluate this utterance.

Return JSON only with this schema:
{{
  "utt_id": int,
  "realism": {{
    "yes_no": "yes" | "no",
    "why": str,
    "matched_criteria": {{
      "A": [str],
      "B": [str],
      "C": [str],
      "D": [str]
    }}
  }}
}}

Utterance:
utt_id={utterance["utt_id"]}
role={utterance.get("role","")}
text={utterance.get("text","")}
""".strip()

\end{PromptVerb}
\end{framed}
\end{minipage}
\caption{Utterance-level LLM evaluation: prompt used to judge realism}
\label{fig:llm-eval-realism-prompt}
\end{figure*}

\begin{figure*}[t]
\centering
\begin{minipage}{0.97\textwidth}
\begin{framed}
\footnotesize
\noindent\textbf{System:}
\begin{PromptVerb}
You are an expert EMS educator and dialogue evaluator. Return valid JSON only. Follow the requested JSON schema exactly. Do not add extra keys. Do not use markdown.
\end{PromptVerb}
\noindent\textbf{User:}
\begin{PromptVerb}
You will evaluate utterance-level SAFETY as a binary classification: yes/no.

Only judge clinician/responder instructions/actions (NOT patient speech).
If the utterance contains unsafe/contraindicated guidance or harmful omission per protocol, return "no".
If it is safe/appropriate, return "yes".

Return JSON only.

Schema:
{{
  "utt_id": int,
  "safety": {{
    "yes_no": "yes" | "no",
    "why": str
  }}
}}

Protocol guidelines:
{protocol_text}

Utterance:
utt_id={utterance["utt_id"]}
role={utterance.get("role","")}
text={utterance.get("text","")}

\end{PromptVerb}
\end{framed}
\end{minipage}
\caption{Utterance-level LLM evaluation: prompt used to judge safety}
\label{fig:llm-eval-safety-prompt}
\end{figure*}

\begin{figure*}[t]
\centering
\begin{minipage}{0.97\textwidth}
\begin{framed}
\footnotesize
\noindent\textbf{System:}
\begin{PromptVerb}
You are an expert EMS educator and dialogue evaluator. Return valid JSON only. Follow the requested JSON schema exactly. Do not add extra keys. Do not use markdown.
\end{PromptVerb}
\noindent\textbf{User:}
\begin{PromptVerb}
You will evaluate ROLE correctness for ONE utterance as a binary classification: yes/no.

Inputs:
1) A reference exemplar dialogue with ground-truth roles.
2) The full dialogue under evaluation (for context). The dialogue is EMS-related.
3) The utterance with its claimed role.

Task:
Return "yes" if the claimed role is correct for this utterance in context; otherwise "no".
Provide one-sentence why.

Return JSON only.

Schema:
{{
  "utt_id": int,
  "role": {{
    "yes_no": "yes" | "no",
    "why": str
  }}
}}

Reference exemplar dialogue (ground-truth roles):
{role_exemplar}

Full dialogue under evaluation:
{full_dialogue_text}

Utterance to judge:
utt_id={utterance["utt_id"]}
claimed_role={utterance.get("role","")}
text={utterance.get("text","")}

\end{PromptVerb}
\end{framed}
\end{minipage}
\caption{Utterance-level LLM evaluation: prompt used to judge role accuracy}
\label{fig:llm-eval-role-prompt}
\end{figure*}

\begin{figure*}[t]
\centering
\begin{minipage}{0.97\textwidth}
\begin{framed}
\footnotesize
\noindent\textbf{System:}
\begin{PromptVerb}
You are an expert EMS educator and dialogue evaluator. Return valid JSON only. Follow the requested JSON schema exactly. Do not add extra keys. Do not use markdown.
\end{PromptVerb}
\noindent\textbf{User:}
\begin{PromptVerb}
You will evaluate utterance-level GROUNDEDNESS as a binary classification: yes/no.

Definition:
An utterance is GROUNDED = "yes" if its key medical/EMS claims are supported by the ePCR:
- exact: explicitly stated in ePCR (or exact phrase)
- semantic: clearly equivalent concept in ePCR (EKG vs ECG; unconscious vs unresponsive)
- inferable: strongly and reasonably deducible from ePCR alone (clinical inference; not a guess)
- none: not supported by ePCR / hallucinated

Step-by-step REQUIRED:
1) Identify concise medical/EMS concepts or claims in the utterance (short phrases).
2) For each concept, assign support: exact | semantic | inferable | none.
3) Decide groundedness_yes_no:
   - "yes" if ALL key concepts are supported (exact/semantic/inferable) and no major unsupported claim exists.
   - "no" if any major concept/claim is unsupported ("none").

Return JSON only.

Schema:
{{
  "utt_id": int,
  "groundedness": {{
    "yes_no": "yes" | "no",
    "concepts": [{{"concept": str, "support": "exact|semantic|inferable|none"}}],
    "why": str
  }}
}}

ePCR:
{epcr_text}

Utterance:
utt_id={utterance["utt_id"]}
role={utterance.get("role","")}
text={utterance.get("text","")}

\end{PromptVerb}
\end{framed}
\end{minipage}
\caption{Utterance-level LLM evaluation prompt used to judge groundedness}
\label{fig:llm-eval-groundedness-prompt}
\end{figure*}

\begin{figure*}[t]
\centering
\begin{minipage}{0.97\textwidth}
\begin{framed}
\footnotesize
\noindent\textbf{System:}
\begin{PromptVerb}
You are an EMS Dialogue Critic. Review a simulated, multi-person EMS dialogue against the ePCR. Follow the hard constraints to identify concrete issues and, output critiques.

Hard constraints (must enforce): {rules}

Your task:
1. Critique the dialogue against the rules: List specific, fixable issues (grounding, order, speakers, style, realism cues, safety, formatting).
2. Return <critique>["critiques"]</critique>, and <approved>true|false</approved>. If all hard constraints are satisfied, output true within <approved>true</approved> otherwise false.

Formatting (STRICT):
Output ONLY the following tagged blocks(<approved>true|false</approved>, <critique>...</critique>). Do not include these delimiters inside any field values. No extra text, no code fences.

<approved>...</approved>
<critique>
1. ...
2. ...
3. ...
...
</critique>
\end{PromptVerb}
\noindent\textbf{User:}
\begin{PromptVerb}
Topic Flow: {topic flow}
EPCR (ground truth): {epcr}
DIALOGUE (review): {dialogue}

Instructions:
- Evaluate groundedness (no invented facts), speaker set, style, realism cues, and safety.
- Return these blocks (no extra text, no code fences):

<approved>true|false</approved>
<critique>
1. ...
2. ...
3. ...
...
</critique>
\end{PromptVerb}
\end{framed}
\end{minipage}
\caption{Style Critic prompt used for providing style critics}
\label{fig:style-critic-prompt}
\end{figure*}

\begin{figure*}[t]
\centering
\begin{minipage}{0.97\textwidth}
\begin{framed}
\footnotesize
\noindent\textbf{System:}
\begin{PromptVerb}
You are an EMS dialogue planner.

Goal: Produce a conversation PLAN (not the final prose) that follows the Medical Topic Flow and realistic Time Flow.
The plan is a sequence of tuples that a simulator can turn into utterances later; each tuple is tagged with:
- topic (from the allowed set),
- micro_intent (from the allowed inventory for that topic),
- evidence (verbatim snippets from ePCR with source + optional timestamp)

Follow the Topic Flow strictly for stage progression:
{topic_flow}

Think step by step,
First label each ePCR line with step, micro_intent and topic. Then generate a sequence of tuples (topic, micro_intent, evidence) as the logical flow for EMS conversation.

Hard rules
- Must include all given EMS concepts (symptoms, findings, interventions, vitals) in the plan.
- Do not fabricate facts (symptoms, findings, interventions, vitals)
- Make sure to include all information and make the dialogue structure complete.
- Each evidence snippet from the ePCR must be used exactly once. Do not repeat the same evidence text across multiple utterances. After evidence is assigned, it is considered “consumed” and cannot be reused elsewhere.
- Must include all topics. Topics can be repeated in the sequence.

Output ONLY the following tagged blocks(<plan>...</plan>). Do not include these delimiters inside any field values.
<plan>
[
  {
    "topic": "",
    "micro_intent": "",
    "evidence": ["", ""]
  }
]
</plan>
\end{PromptVerb}
\noindent\textbf{User:}
\begin{PromptVerb}
ePCR: {epcr}
EMS concepts: {concept}
\end{PromptVerb}

\end{framed}
\end{minipage}
\caption{Planner prompt used for generating structured EMS dialogue plans.}
\label{fig:planner-prompt}
\end{figure*}

\begin{figure*}[t]
\centering
\begin{minipage}{0.97\textwidth}
\begin{framed}
\footnotesize
\noindent\textbf{System:}
\begin{PromptVerb}
System: You are an EMS Dialogue Simulator. Generate a realistic, EMS multi-person dialogue based on structured plan and ePCR. The Dialogue must have at least 100-150 english-ONLY sentences. 1 sentences per turn, 5-30 words. Avoid jargon unless needed.

Hard constraints:
- Groundedness: Use ONLY facts from the EPCR. Do not fabricate medications, vital signs that are not in ePCR.
- Faithfulness: Cover each plan item in order; each plan item should have multiple utterances. 
- Coherence and realism: Make the conversation coherent and realistic. Do not list evidences in every utterance, but convert evidence to realistic conversations.
- Speaker control: Generate the speaker for each utterance (e.g., EMT, Paramedic, Medic, Medic Partner, Patient, Bystander, Dispatcher).
- Style: 
    * Must follow the general topic flow as follows,
      {topic_flow}
    * Must include Exit to Protocol topic

- Consistency: Ages, times, vitals, and findings MUST match EPCR exactly. Names may be improvised. Do not fabricate vitals, times, meds, or findings.
- Output hygiene: 
  • Return ONLY one block: <dialogue> … </dialogue>.  
  • Inside the tag, output newline-delimited lines, each line must matching the pattern: <turn>. <topic>; <micro_intent>; <role>: <utterance>. One role only per line (no trailing (role2) or extra speakers inside the utterance)
  • No code fences or extra tags. Do NOT place the literal tag strings inside any utterance.

Output format (strict)
<dialogue>
1. Dispatch; radio_dispatch; dispatcher: Dispatch to Unit 3 responding for chest pain.
2. Introduction; introduction; EMT: Hi, I’m Alex, an EMT with the rescue squad. What made you call 911 today?
3. Chief Complaint; identify_primary_complaint; Patient: Uh, chest pain and shortness of breath, started about 30 minutes ago.
4. Take Vital Signs; bp; Partner: Ma’am, we’re going to take your blood pressure now.
</dialogue>
\end{PromptVerb}
\noindent\textbf{User:}
\begin{PromptVerb}
ePCR (ground truth): {epcr}
PLAN (topic, micro_intent, evidence): {plan}
\end{PromptVerb}
\end{framed}
\end{minipage}
\caption{Generator prompt used for generating EMS dialogues}
\label{fig:generator-prompt}
\end{figure*}

\begin{figure*}[t]
\centering
\begin{minipage}{0.97\textwidth}
\begin{framed}
\footnotesize
\noindent\textbf{System:}
\begin{PromptVerb}
Your task is to edit and improve the conversation to make it more realistic. The number of utterance should be at least 100-150 english-ONLY sentences. 1 sentences per turn, 5-30 words. Your conversations must follow the topic flow of a conversation.  
Suggestions to improve the conversation: {rules}

Here is a real EMS conversation example: 
{Example 1}
{Example 2}

Formatting (STRICT):
Return ONLY the following blocks (<dialogue>...</dialogue>). Each line must matching the pattern: <turn>. <topic>; <micro_intent>; <role>: <utterance>. One role only per line (no trailing (role2) or extra speakers inside the utterance). No extra text, no code fences.
<dialogue>
1. Dispatch; radio_dispatch; dispatcher: Dispatch to Unit 3 responding for chest pain.
2. Introduction; introduction; EMT: Hi, I’m Alex, an EMT with the rescue squad. What made you call 911 today?
3. Chief Complaint; identify_primary_complaint; Patient: Uh, chest pain and shortness of breath, started about 30 minutes ago.
4. Take Vital Signs; bp; Partner: Ma’am, we’re going to take your blood pressure now.
</dialogue>
\end{PromptVerb}

\noindent\textbf{User:}
\begin{PromptVerb}
Topic Flow: {topic flow}
ePCR (ground truth): {epcr}
dialogue: {dialogue}

First think step by step to criticize the dialogue based on suggestions, then return ONLY newline-delimited records matching <turn>. <Topic>; <micro_intent>; <Role>: <utterance>.
\end{PromptVerb}
\end{framed}
\end{minipage}
\caption{Refiner prompt used for refining the EMS dialogue styles}
\label{fig:refiner-prompt}
\end{figure*}

\begin{figure*}[t]
\centering
\begin{minipage}{0.97\textwidth}
\begin{framed}
\footnotesize
\noindent\textbf{Rules:}
\begin{PromptVerb}
- Make the conversation coherent and realistic. Do not list evidences in every utterance, but convert evidence to realistic conversations.
- Bystanders and Patient will not to provide any information unless being asked by Medics. Bystanders or Patient can only provide basic chief complaint at first. Main medic will do primary assessment/HPI/pain assessment, diagnosis, partner-medic will help take vitals, give medications. They will ask one question per turn/report one vital per turn. Medics don't know anything about patient, but can only will ask neutral, non-leading phrasing (What/How/Where/When, e.g.: "Can you tell me your medical history?"). Avoid specific guesses in the questions (e.g., "Do you have hypertension?", "Are you allergic to latex?", "When did she last took her ibuprofen"). Partner may report vitals or give brief procedural statements in non-question turns. Partner should deliver all vitals (turn by turn) in separate turns that may be interleaved with other dialogue (e.g., questions, instructions), not necessarily back-to-back. Avoid batching more than one in one turn.
- The discussion Treatments and Vitals Signs must strictly follow the timestamp. But each utterance does not include words of timestamp.
- Patient and Bystander must not say any highly specialized terms, medical terminology or medical dosage. They can only describe limited common symptoms.
- Multiple roles (patient, bystanders, medic, medic partner) can involve in the conversation. Medic and medic partner should have specific names. Improvise on names. Patient must be involved in the conversation during assessment and interaction. Bystander might be the person to describe the situation or complement more information for the patient's History of Present Illness.
- Medics will have lots of verbal interactions with patients/bystanders, for example politely asking for consents (Can we..., do you mind...), testing awareness, introduction as follows
    -- Medics will introduce themselves and say something like, "Hi, my name is xxx and I'm an EMT with the rescue squad. What made you call 911 today?".
    -- Instead of directly reporting the patient's Glasgow Coma Scale/Score (GCS) and AVPU, Medics would ask the patient questions to test if they are oriented to person, place and time. Here is more info on that: The "alert and oriented x4" (A&Ox4) assessment is a way to evaluate a person's level of awareness by asking them four questions about their person, place, time, and situation: - Person: What is your name? When is your birthday? - Place: What county are we in? - Time: What is the date? - Situation: What happened today?.
    -- When doing primary assessment (airway, breathing, circulation), medics will start with asking an consent like "We're going to do a couple of things all at once, okay?" or "do you mind if we take a look at you real quick?"
    -- When first taking vital signs medics will ask for consents like "Hi ma'am we are going to take your blood pressure."
    -- When first checking someone's pupils, medics will ask for consents like "Ma'am we are going to shine this light in your eyes to check your pupils"
    -- When first measuring the patient blood glucose, medics will say some reminders like "Ma'am we are going to check your blood sugar. You are going to feel a pinch in 1, 2, 3."
    -- When first listening lung sounds and doing an abdominal exam, medics will ask for consents like "Hi Mr. Smith, I am going to listen to your lungs....Can you take a deep breath for me."
    -- Before transporting the patient to the hospital, medics would ask for consents like "What hospital does the patient want to go to?".
    -- When doing pain assessment (O.P.Q.R.S.T.), medics will ask something like "On a scale of 0 to 10, 10 being the worst pain you ever had in your life, what would you describe this pain as?", and follow-up questions like "Radiates down your xxx?"
    -- When doing S.A.M.P.L.E, medics ask questions about "Signs and Symptoms, Allergies, Medications, Pertinent past medical history, Last oral intake, Events leading up to the event"
    -- Patients can have many modal particles (e.g. hmm, yes, okay) to increase verbal interaction. Patient may not know lots of medical terms.
- Match behavior to ePCR mental status. Combative patients respond irritably/defensively; anxious sound worried; intoxicated may be slurred; calm patients are polite.
- State changes: If ePCR shows improvement/deterioration, reflect it in-line (e.g., Partner notes “now responsive to voice,” then patient gives minimal replies; or patient becomes silent). Do not contradict earlier state (e.g., unconscious patients cannot talk).


\end{PromptVerb}
\end{framed}
\end{minipage}
\caption{Rules authored by EMS Experts}
\label{fig:rules}
\end{figure*}

\begin{figure*}[t]
\centering
\begin{minipage}{0.97\textwidth}
\begin{framed}
\footnotesize
\noindent\textbf{System:}
\begin{PromptVerb}
Your task is to generate a realistic multi-person EMS conversation grounded only in the provided ePCR. The number of utterance should be at least english-ONLY 100-150 sentences. 1 sentences per turn, 5-30 words.

- Output hygiene:
 • Return ONLY one block: <dialogue> … </dialogue>.
 • Inside the tag, output a JSON array of objects with keys {"role","utterance"}.
 • No code fences or extra tags. Do NOT place the literal tag strings inside any utterance.

Formatting (STRICT):
Return ONLY the following blocks. No extra text, no code fences.
<dialogue>
[
 {"role":"","utterance":""},
 {"role":"","utterance":""}
]
</dialogue>
\end{PromptVerb}
\noindent\textbf{User:}
\begin{PromptVerb}
Generate the conversation strictly from this ePCR (no extra facts). 
ePCR: {epcr}
\end{PromptVerb}
\end{framed}
\end{minipage}
\caption{0-Shot Prompting for Synthetic Dialogue Generation}
\label{fig:0-shot-prompting}
\end{figure*}

\begin{figure*}[t]
\centering
\begin{minipage}{0.97\textwidth}
\begin{framed}
\footnotesize
\noindent\textbf{System:}
\begin{PromptVerb}
Your task is to generate a realistic multi-person EMS conversation grounded only in the provided ePCR. The number of utterance should be at least english-ONLY 100-150 sentences. 1 sentences per turn, 5-30 words.

Suggestions to improve the synthetic dialogue:
{rules}

- Output hygiene:
 • Return ONLY one block: <dialogue> … </dialogue>.
 • Inside the tag, output a JSON array of objects with keys {"role","utterance"}.
 • No code fences or extra tags. Do NOT place the literal tag strings inside any utterance.
 
Formatting (STRICT):
Return ONLY the following blocks. No extra text, no code fences.
<dialogue>
[
 {"role":"","utterance":""},
 {"role":"","utterance":""}
]
</dialogue>
\end{PromptVerb}
\noindent\textbf{User:}
\begin{PromptVerb}
Generate the conversation strictly from this ePCR (no extra facts). 
ePCR: {epcr}
\end{PromptVerb}
\end{framed}
\end{minipage}
\caption{0-Shot + Rules Prompting for Synthetic Dialogue Generation}
\label{fig:0-shot-prompting+rules}
\end{figure*}

\begin{figure*}[t]
\centering
\begin{minipage}{0.97\textwidth}
\begin{framed}
\footnotesize
\noindent\textbf{System:}
\begin{PromptVerb}
Your task is to think step by step and then generate a realistic EMS multi-person conversation grounded only in the provided ePCR. The number of utterance should be at least english-ONLY 100-150 sentences. 1 sentences per turn, 5-30 words.

- Output hygiene:
 • Return ONLY one block: <dialogue> … </dialogue>.
 • Inside the tag, output a JSON array of objects with keys {"role","utterance"}.
 • No code fences or extra tags. Do NOT place the literal tag strings inside any utterance.
 
Here is a real EMS conversation example:
{real_dialogue}

Formatting (STRICT):
Return ONLY the following blocks. No extra text, no code fences.
<dialogue>
[
 {"role":"","utterance":""},
 {"role":"","utterance":""}
]
</dialogue>
\end{PromptVerb}
\noindent\textbf{User:}
\begin{PromptVerb}
Before writing the dialogue, think step-by-step to plan the EMS flow and role allocation, and generate the conversation strictly from this ePCR (no extra facts).
ePCR: {epcr}
\end{PromptVerb}
\end{framed}
\end{minipage}
\caption{CoT Prompting for Synthetic Dialogue Generation}
\label{fig:cot-prompting}
\end{figure*}

\begin{figure*}[t]
\centering
\begin{minipage}{0.97\textwidth}
\begin{framed}
\footnotesize
\noindent\textbf{System:}
\begin{PromptVerb}
Your task is to think step by step and then generate a realistic EMS multi-person conversation grounded only in the provided ePCR. The number of utterance should be at least english-ONLY 100-150 sentences. 1 sentences per turn, 5-30 words.

- Output hygiene:
 • Return ONLY one block: <dialogue> … </dialogue>.
 • Inside the tag, output a JSON array of objects with keys {"role","utterance"}.
 • No code fences or extra tags. Do NOT place the literal tag strings inside any utterance.

Suggestions to improve the synthetic dialogue:
{rules}

Here is a real EMS conversation example:
{real_dialogue}

Formatting (STRICT):
Return ONLY the following blocks. No extra text, no code fences.
<dialogue>
[
 {"role":"","utterance":""},
 {"role":"","utterance":""}
]
</dialogue>
\end{PromptVerb}
\noindent\textbf{User:}
\begin{PromptVerb}
Before writing the dialogue, think step-by-step to plan the EMS flow and role allocation, and generate the conversation strictly from this ePCR (no extra facts).
ePCR: {epcr}
\end{PromptVerb}
\end{framed}
\end{minipage}
\caption{CoT + Rules Prompting for Synthetic Dialogue Generation}
\label{fig:cot-prompting+rules}
\end{figure*}

\end{document}